\documentclass[
]{ceurart}

\usepackage{listings}
\usepackage{soul}

\usepackage{algorithmic}
\usepackage{algorithm}

\usepackage{pifont}%
\newcommand{\cmark}{\ding{51}}%
\newcommand{\xmark}{\ding{55}}%

\usepackage{url}

\usepackage{subcaption}

\usepackage{amsthm}
\newtheorem{theorem}{Theorem}[section]
\newtheorem{lemma}[theorem]{Lemma}

\begin{document}

\copyrightyear{2025}
\copyrightclause{Copyright for this paper by its authors.
  Use permitted under Creative Commons License Attribution 4.0
  International (CC BY 4.0).}

\conference{4DMR@IJCAI25: International IJCAI Workshop on 1st Challenge and Workshop for 4D Micro-Expression Recognition for Mind Reading, August 29, 2025, Guangzhou, China.}

\title{Inference-Time Gaze Refinement for Micro-Expression Recognition: Enhancing Event-Based Eye Tracking with Motion-Aware Post-Processing}


\author[1]{Nuwan Bandara}[%
orcid=0009-0002-2509-5117,
email=pmnsbandara@smu.edu.sg,
url=https://www.nuwanbandara.com/,
]
\address[1]{School of Computing and Information Systems, Singapore Management University, Singapore}

\author[1]{Thivya Kandappu}[%
orcid=0000-0002-4279-2830,
email=thivyak@smu.edu.sg,
url=https://www.thivyak.info/,
]

\author[1]{Archan Misra}[%
orcid=0000-0003-1212-1769,
email=archanm@smu.edu.sg,
url=https://sites.google.com/view/archan-misra,
]


\begin{abstract}
  Event-based eye tracking holds significant promise for fine-grained cognitive state inference, offering high temporal resolution and robustness to motion artifacts, critical features for decoding subtle mental states such as attention, confusion, or fatigue. In this work, we introduce a model-agnostic, inference-time refinement framework designed to enhance the output of existing event-based gaze estimation models without modifying their architecture or requiring retraining. Our method comprises two key post-processing modules: (i) Motion-Aware Median Filtering, which suppresses blink-induced spikes while preserving natural gaze dynamics, and (ii) Optical Flow-Based Local Refinement, which aligns gaze predictions with cumulative event motion to reduce spatial jitter and temporal discontinuities. To complement traditional spatial accuracy metrics, we propose a novel Jitter Metric that captures the temporal smoothness of predicted gaze trajectories based on velocity regularity and local signal complexity. Together, these contributions significantly improve the consistency of event-based gaze signals, making them better suited for downstream tasks such as micro-expression analysis and mind-state decoding. Our results demonstrate consistent improvements across multiple baseline models on controlled datasets, laying the groundwork for future integration with multimodal affect recognition systems in real-world environments. Our code implementations can be found at \url{https://github.com/eye-tracking-for-physiological-sensing/EyeLoRiN}.
\end{abstract}


\maketitle

\section{Introduction}
\label{sec:intro}

Micro-expressions, brief, involuntary facial expressions that reveal genuine emotional states, have gained significant attention as a non-verbal cue for "mind reading", i.e., inferring internal cognitive or affective states. Unlike overt expressions, micro-expressions are transient (typically lasting less than 0.5 seconds) and often escape conscious control, making them a powerful signal for detecting concealed emotions or subtle shifts in mental state~\cite{ekman2004emotions, yan2014casme}. Applications of micro-expression analysis span diverse fields, including high-stakes security screening~\cite{ekman2009telling}, mental health assessment~\cite{wezowski2025open}, affective computing~\cite{wang2020micro}, and adaptive human-computer interaction~\cite{alipour2023framework}. With the growing demand for systems that can sense and respond to users' cognitive-affective dynamics in real-time, there is increasing interest in leveraging micro-expressions for fine-grained, moment-to-moment mind reading.

Among the various components of micro-expression analysis, ocular behavior, including pupil dilation, blink rate, and eye movement dynamics, has emerged as a particularly informative subregion. Eye movements, in particular, offer a window into the user's attentional focus, cognitive load, emotional arousal, and even motivational state. Prior studies have linked gaze dynamics to transient mental states such as confusion, boredom, interest, and fatigue~\cite{bixler2015automatic, eckstein2017beyond, prasse2024improving}. Rapid shifts in gaze (e.g., saccades), subtle fixations, and jitter-like microsaccades can serve as physiological markers for processes such as attentional lapses or spontaneous engagement. As such, high-fidelity eye tracking forms a crucial input for real-time micro-expression-based inference systems that aim to detect and respond to users' unspoken mental states.

However, traditional camera-based eye tracking systems face well-known limitations in capturing these micro-level dynamics. Frame-based approaches typically operate at 30 – 1000 Hz and struggle with motion blur during rapid eye movements or transient behaviors such as blinking. In contrast, event-based vision sensors capture pixel-level brightness changes asynchronously with microsecond latency, yielding sparse but high-temporal-resolution data streams. These characteristics make event-based sensors ideally suited for fine-grained eye tracking, especially in contexts where temporal fidelity, motion robustness, and low-latency processing are critical.

Despite these advantages, effective utilization of event data in eye tracking remains a challenge. The sparse and asynchronous nature of the data presents difficulties in processing and interpreting, requiring specialized models that can handle both the temporal and spatial dimensions of eye movements~\cite{chen2025eventvision_event}. Spatio-temporal models, such as Change-Based ConvLSTM~\cite{chen20233et}, graph-based event representations~\cite{bandara2024eyegraph}, and event binning methods~\cite{pei2024lightweight}, have emerged as promising approaches. These models aim to capture the dynamic and continuous nature of eye movements by encoding both spatial and temporal information from the event streams. This is particularly important because eye movements are inherently continuous, both in terms of space and time, and spatio-temporal models attempt to leverage these properties to improve gaze estimation accuracy.

However, while these models have shown promise, they suffer from several limitations. A key challenge in event-based eye tracking is the handling of blink artifacts, which cause interruptions in the event data and lead to erroneous gaze predictions~\cite{grootjen2023highlighting}. Another limitation is the temporal inconsistency often observed in the predictions, as eye movements are physiologically continuous and models sometimes fail to enforce this temporal smoothness, leading to abrupt gaze shifts that undermine tracking stability~\cite{allopenna1998tracking}. Additionally, existing models often fail to fully leverage local event distributions, resulting in misaligned gaze predictions. These challenges, coupled with the inherent label sparsity of event datasets, make it difficult to develop a universally robust event-based tracking system.

To address these challenges, we propose a model-agnostic inference-time post-processing and local refinement framework to enhance the accuracy and robustness of event-based eye tracking. Our approach targets the shortcomings of existing spatio-temporal models by introducing lightweight, post-processing techniques that can be integrated with any model without requiring retraining or architectural changes. This makes our method flexible and easily applicable to a wide range of existing models. The post-processing framework consists of two key components: (i) motion-aware median filtering, which enforces temporal smoothness by taking advantage of the continuous nature of eye movements, and (ii) optical flow-based local refinement, which improves spatial consistency by aligning gaze predictions with dominant motion patterns in the local event neighborhood. These refinements not only mitigate blinking artifacts but also ensure that gaze predictions remain temporally continuous and spatially accurate, even in the presence of rapid eye movements or motion artifacts.

By incorporating these post-processing and refinement techniques, our approach improves the overall performance and robustness of event-based eye tracking. This makes it particularly valuable in real-world applications where traditional models may fail due to the challenges posed by low-light environments, high-speed motion, or intermittent artifacts like blinks. Furthermore, because our framework is model-agnostic, it can be applied to any existing event-based eye-tracking model, offering a significant boost to accuracy and stability without requiring changes to the core model.

In this paper, we make the following key contributions:
\begin{itemize}
    \item \textbf{Model-Agnostic Post-Processing:} We propose an inference-time refinement approach that enhances existing event-based pupil estimation models without modifying their architectures or requiring retraining through (1) \textbf{Motion-Aware Median Filtering:}  a median filtering technique that incorporates motion-awareness to preserve temporal continuity in gaze predictions and mitigate blinking-induced errors, and (2) \textbf{Optical Flow-Based Smooth Refinement:} a local refinement strategy leveraging optical flow to ensure that predicted gaze positions align with the cumulative local event motion, reducing spatial inconsistencies in tracking results.
    \item \textbf{Jitter Metric:} We propose a complementary metric for pupil tracking task to specifically evaluate the temporal smooth continuity of the predictions with respect to true targets based on (1) global statistical distribution of pupil velocities, addressed via comparative velocity entropy, and (2) local fine-grained frequency content of pupil velocities, addressed via spectral arc length-guided spectral entropy.
    \item \textbf{Empirical Validation and Performance Gains:} Through extensive experiments, we demonstrate that our proposed methods significantly enhance the robustness and accuracy of state-of-the-art event-based eye-tracking models across diverse conditions. 
\end{itemize}



\section{Related Work}

\subsection{Micro-expression Recognition}
Micro-expressions are rapid, involuntary facial movements that reveal transient emotional states often concealed from conscious awareness. Due to their subtlety and brevity, automatic recognition of micro-expressions remains a challenging task that has garnered significant interest within the computer vision and affective computing communities.

Early approaches predominantly relied on handcrafted spatio-temporal features extracted from high-frame-rate facial video sequences, such as Local Binary Patterns on Three Orthogonal Planes (LBP-TOP)~\cite{guo2019extended}, optical flow-based descriptors ~\cite{verburg2019micro}, and optical strain~\cite{liong2015subtle}, to capture nuanced facial muscle dynamics. While these methods laid the foundation for subsequent advances, they were often limited by their dependence on feature engineering and sensitivity to noise. More recently, deep learning architectures, including 3D Convolutional Neural Networks~\cite{xia2019spatiotemporal} and Long Short-Term Memory (LSTM) networks~\cite{bai2020investigating}, have been employed to automatically learn hierarchical representations from high-frame-rate facial video sequences, significantly improving recognition accuracy. Attention mechanisms have further enhanced model capacity by focusing on discriminative spatial and temporal regions~\cite{wang2023micro}. The availability of specialized datasets such as CASME~\cite{yan2013casme}, CASME II~\cite{yan2014casme}, SMIC~\cite{li2013spontaneous}, and SAMM~\cite{davison2016samm}, has been foundational, providing high-frame-rate facial videos annotated with micro-expression labels under controlled environments.

While facial cues remain the primary modality for micro-expression analysis, recent studies underscore the complementary value of ocular signals, including pupil dynamics and saccadic eye movements, as indicators of cognitive and affective states~\cite{hess1960pupil, partala2003pupil}. Integrating eye-tracking data can enhance the robustness and granularity of emotion recognition, particularly in applications such as deception detection, clinical assessment of affective disorders, and adaptive human-computer interaction systems that respond to user engagement and mental workload.

In this paper, we specifically address the challenge of fine-grained eye tracking, an important yet often overlooked component of micro-expression analysis, and propose novel techniques to improve its accuracy and temporal consistency, thereby enhancing the reliability of eye-based cognitive and affective state inference.

\subsection{Event-based Pupil and Gaze Tracking}

Event-based cameras, offer a fundamentally different sensing paradigm compared to conventional frame-based cameras. By asynchronously recording pixel-level brightness changes with microsecond temporal resolution and an extremely high dynamic range ($120 dB$) while consuming minimal power ($mW $ level), these sensors are uniquely suited for applications requiring precise, low-latency motion tracking~\cite{gallego2020event}. Consequently, the eye tracking community has recently begun to explore event-based approaches for pupil and gaze tracking, motivated by the limitations of traditional RGB and infrared systems in terms of temporal resolution and power efficiency~\cite{bandara2024eyegraph}.
Current research on event-based eye tracking can be broadly categorized into two main approaches:

\textbf{Hybrid event-RGB:} Several works have proposed combining event data with RGB frames to leverage the spatial resolution and structural detail of conventional cameras alongside the temporal advantages of event sensors. The approaches presented in~\cite{angelopoulos2021event} exemplify this direction by using RGB frames for initial pupil detection, with event streams employed to refine temporal tracking. While effective in improving robustness, such hybrid methods are inherently constrained by the frame rate of the RGB sensor, typically in the order of tens of milliseconds, which limits the full exploitation of the asynchronous, high-frequency event data. Moreover, these methods rely on RGB imagery and thus may suffer from environmental limitations such as variable lighting conditions, which event sensors are inherently more resilient to.

\textbf{Event-Only Tracking:}
More recent work focuses exclusively on event streams, aiming to fully harness the unique properties of event cameras for pupil and gaze estimation. The event-only approaches~\cite{sen2024eyetraes, bandara2024eyegraph, wang2024event}, which aggregate events into either 2D or 3D representations and get inferred through either neural networks or traditional computer vision algorithms and thus, occasionally suffer from label sparsity and inefficient representations.

Bandara et al.~\cite{bandara2024eyegraph} proposed EyeGraph, a novel approach that constructs spatiotemporal graphs from event data to represent pupil contours, and addressed the issue of label sparsity by proposing an unsupervised graph-based clustering approach to spatially localise the pupil in a 3D event volume. In contrast, Sen et al.~\cite{sen2024eyetraes} presented EyeTrAES, an event-based adaptive slicing mechanism that adaptively adjusts the volume of the event accumulation based on the underlying eye motion.

Despite these advances, event-only pupil tracking faces several open challenges. Designing representations that preserve the high temporal fidelity of events without overwhelming computational resources remains an active research area. Moreover, the limited availability of synchronized ground truth data hinders large-scale supervised training. Event sensors also produce noise from environmental artifacts such as illumination flicker or head movement, necessitating robust filtering and model designs~\cite{gallego2020event}.


\subsection{Spatio-Temporal Processing of Events}
Event-based eye tracking requires models that effectively capture the sparse, asynchronous nature of event streams while preserving high temporal resolution~\cite{gallego2020event}. Unlike frame-based methods, event-based vision demands spatio-temporal representations that can model continuous eye movements with precision.
Recent prior works have proposed several spatio-temporal models-based event processing to capture the temporal evolution of the events. In this paper, we consider three classes of such models: (a) ConvGRU and ConvLSTM, (b) graph-based representations, and (c) event-binning methods.

ConvLSTM and ConvGRU networks~\cite{chen20233et}, originally designed for dense sequential data, struggle with the sparsity of event streams. Change-Based-convLSTM (CB-convLSTM) mitigates this by leveraging change-based updates, focusing on local event dynamics rather than static frames \cite{chen20233et}. This improves tracking accuracy by ensuring temporal continuity while maintaining event efficiency. Graph-based models encode events as nodes with spatio-temporal edges, preserving fine-grained motion patterns \cite{bandara2024eyegraph}. These approaches enhance tracking by leveraging local event dependencies, making them effective for eye movement estimation. Event binning methods aggregate events over predefined intervals to create structured inputs for deep temporal networks. While simple and efficient, methods like causal event volume binning strike a balance between temporal continuity and real-time feasibility \cite{pei2024lightweight}.

While these models enhance spatio-temporal representation, they often produce temporally inconsistent or spatially misaligned predictions. Our proposed model-agnostic inference-time refinement improves accuracy by enforcing temporal smoothness and aligning predictions with local optical flow. This enhancement operates independently of the underlying spatio-temporal model, making event-based eye tracking more robust across different architectures.

\section{Motivation}
\label{sec: motivation}

\subsection{Inference-time Post-processing}
Even though the current event-based pupil tracking models attempt to incorporate both spatial and temporal learning blocks within their pipelines, the empirical results show that they still suffer from poor performance when handling blink artifacts. Further, even though the pupil movements are physiologically continuous and bounded, these models fail to strictly enforce this rule within the learning pipeline, leading to unstable pupil trajectories at the inference time. To address these limitations, here we propose a motion-aware median filtering technique as a post-processing method which penalizes trajectory outliers, either due to blinking or tracking instability, based on an adaptive motion profiling mechanism and thereby, achieves a stable pupil trajectory. Additionally, given that the existing models often tend to prioritize global perceptible eye morphology compared to local heuristics, which are deduced through event distributions, the pupil trajectory predictions often present an unpreventable offset. To circumvent this issue, we propose to utilize the optical flow around the original predictions such that if the optical flow at the original prediction does not align with the local optical flow, then an offset is assumed and subsequently, corrected by shifting the prediction by a small defined margin. It is to be noted that both of these proposed techniques work in model-agnostic fashion and thus, can be flexibly applied to any existing model without them being re-trained or modified.        



\subsection{Jitter Metric}

Existing metrics for evaluating the event-based pupil tracking performance such as p-accuracy and pixel distances only capture the per-sample positional accuracy of the methods and thereby neglect an essential aspect in pupil tracking: temporal smooth-continuity, which is critical for the stable and user-ergonomic performance in many downstream applications such as foveated rendering, gaze-based human computer interaction, user authentication, and affective-cognitive modelling~\cite{majaranta2014eye, sen2024eyetraes}. In addition, unlike gaze which exhibits both smooth and rapid transitions which sometimes can be almost non-continuous with a significantly higher angle acceleration~\cite{angelopoulos2021event}, the pupil movements are bounded and continuous in nature~\cite{bandara2024eyegraph} which further validates the need for an explicit evaluation metric for temporal smooth-continuity of pupil tracking. To this end, we propose a velocity-based metric which considers and weights both (1) global statistical distribution of velocities via Kullback-Leibler divergence and (2) local velocity jaggedness via spectral arc length-guided spectral entropy. Further, it is to be noted that our metric is comparative between the predicted and true pupil trajectories while being lesser susceptible to measurement or prediction noise in contrast to jerk-based smoothness scores. We theoretically and empirically show the efficacy of the proposed metric in the context of pupil tracking.       

\section{Inference-time Post-processing}

\begin{figure}[!h]
    \centering
    \begin{subfigure}[t]{0.24\linewidth}
        \centering
        \includegraphics[width=\linewidth]{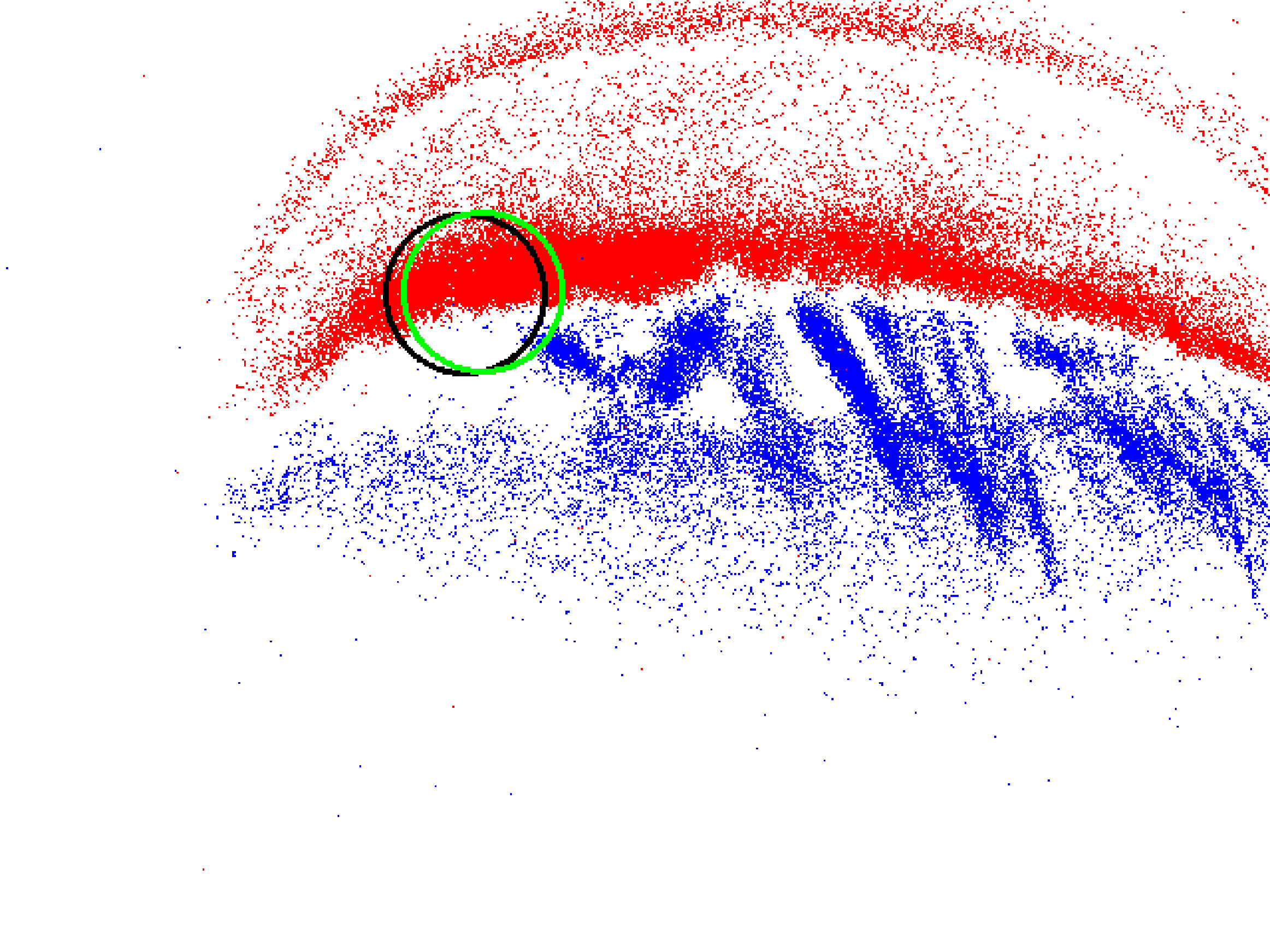}
        \caption{}
        \label{blink_1}
    \end{subfigure}
    \hfill
    \begin{subfigure}[t]{0.24\linewidth}
        \centering
        \includegraphics[width=\linewidth]{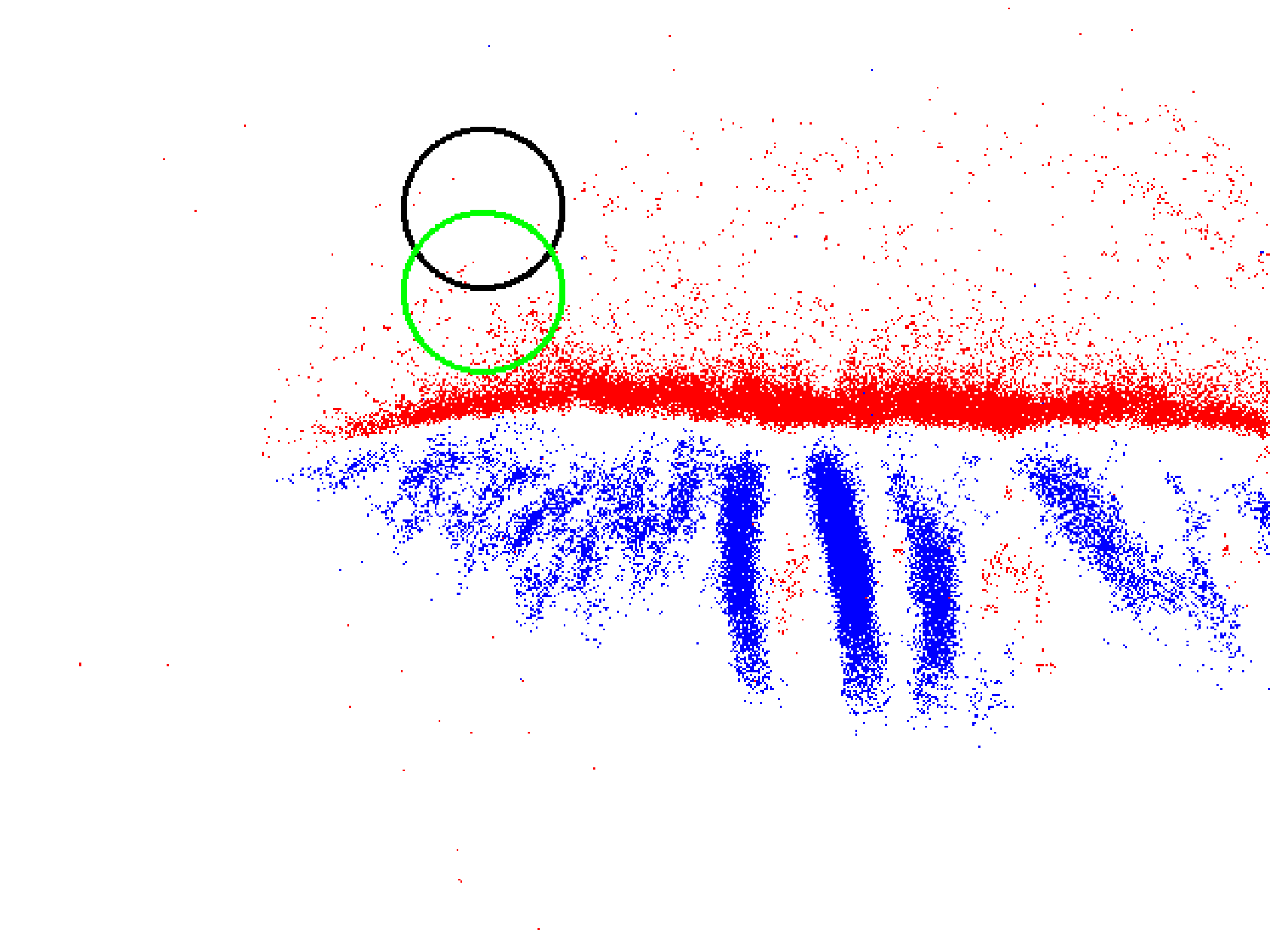}
        \caption{}
        \label{blink_2}
    \end{subfigure}
        \hfill
    \begin{subfigure}[t]{0.24\linewidth}
        \centering
        \includegraphics[width=\linewidth]{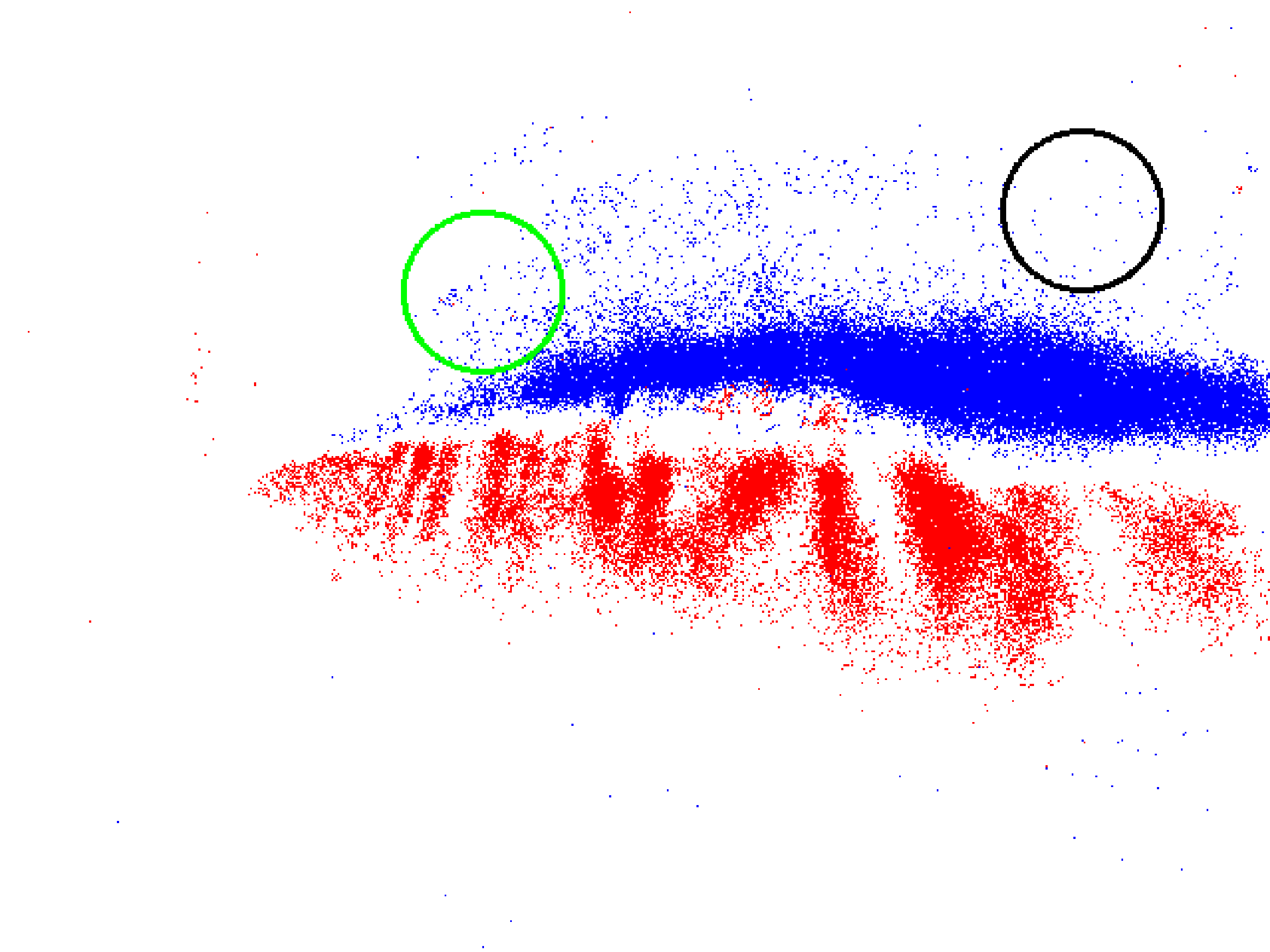}
        \caption{}
        \label{blink_3}
    \end{subfigure}
    \hfill
    \begin{subfigure}[t]{0.24\linewidth}
        \centering
        \includegraphics[width=\linewidth]{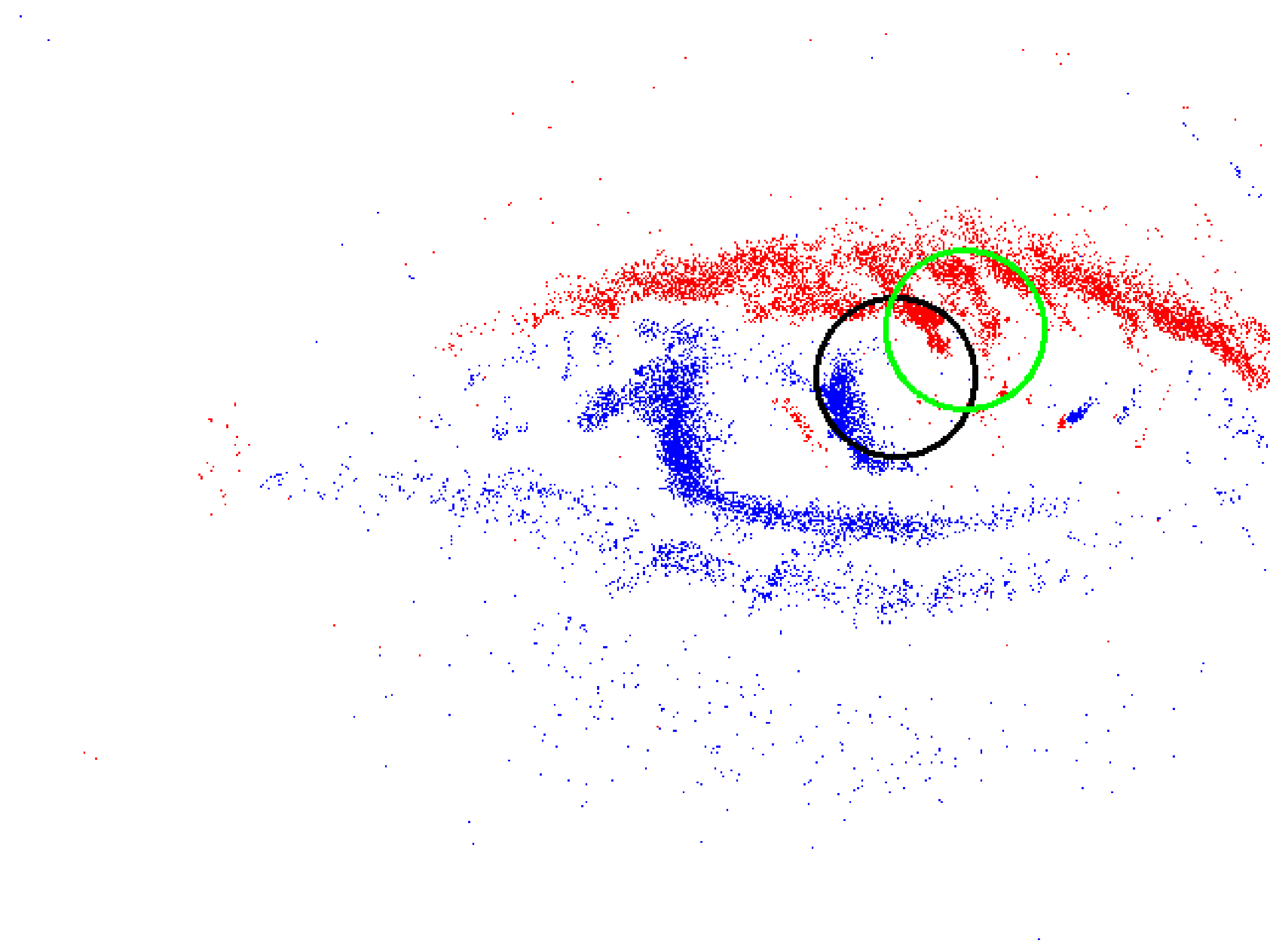}
        \caption{}
        \label{blink_4}
    \end{subfigure}
            \hfill
    \begin{subfigure}[t]{0.5\linewidth}
        \centering
        \includegraphics[width=\linewidth]{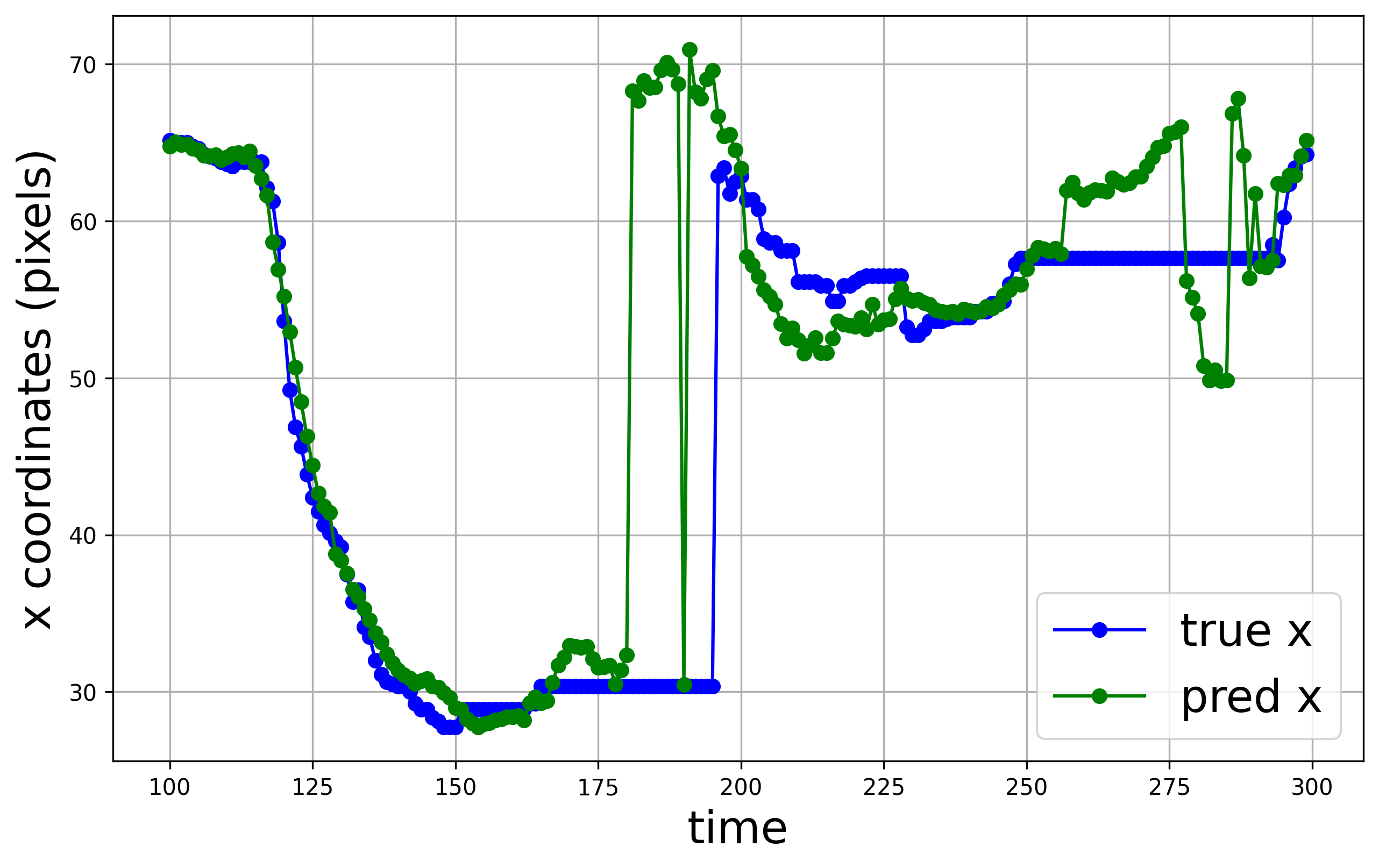}
        \caption{}
        \label{blink_5}
    \end{subfigure}
    \caption{Empirical evidence showing that the blinking artifacts significantly deteriorate the prediction performance. Figure~\ref{blink_1} to~\ref{blink_4} present a near-consecutive scenario where the blinking artifacts (seemingly peaked in Fig.~\ref{blink_3}) occur in which the ground truth is depicted in green while the predictions from~\cite{pei2024lightweight} are shown in black. Both ground truth and predictions are presented with a 5-pixel-radius tolerance circles. Figure~\ref{blink_5} presents the global temporal variation of x coordinates around the presented blinking case. }
    \label{fig: blink_artifacts}
\end{figure}

To address the shortcomings of existing methods in the inference stage, we propose to add two light-weight post-processing techniques specifically targeting the following limitations: (1) motion-aware median filtering (algorithm~\ref{algo:median}) to (a) ensure the temporal consistency of the predictions since the eye movements are physiologically bound to be continuous in spatial domain~\cite{bandara2024eyegraph} and (b) reduce the blinking artifacts and (2) optical flow estimation in the local spatial neighbourhood (algorithm~\ref{algo:ofe}) to smoothly shift the original predictions if the flow vector at the original prediction is unaligned with the cumulative local neighbourhood flow direction. Synoptically, (1) is motivated by our empirical observations which convey the abundance of blinking artifacts within the predictions of the existing models (as shown in Fig.~\ref{fig: blink_artifacts}), whereas (2) is specifically inspired by our observations which hint that the original predictions tend to occupy a negligence towards the event motion flow in the local neighbourhood, suggesting a lack of attention to the local event distribution in the original models. 

More descriptively, in motion-aware filtering as shown in Algorithm~\ref{algo:median}, we first estimate the local motion variance in temporal dimension (i.e., within a set time window) using a set of alternative methods including $0^{th}$ to $2^{nd}$ order kinetics, covariance and frequency (of which the equations are defined in Equations below), and subsequently, assign a median-based adaptive filter windows for each set time windows such that the kernel size for median filtering is adaptive and appropriate to the variability of the background pupil movement while also ensuring the temporal consistency.

With respect to the local motion variance calculation, more mathematically, if the predicted pupil trajectory is \( \mathbf{p}(t) = \begin{bmatrix} x(t) \\ y(t) \end{bmatrix} \) for discrete time index \( t \), then, $0^{th}$ kinetic-based local motion variance estimation could be derived as $\bar{d}_t = \frac{1}{w} \sum_{i=-w/2}^{w/2} \|\ d(t + i)\|$ whereas $d_t = \sqrt{(x_t-x_{t-1})^2+(y_t-y_{t-1})^2}$ and $w$ being the window size. In extension, velocity is the first derivative of position: $\mathbf{v}(t) = \frac{d\mathbf{p}(t)}{dt}$ and thus, the smoothened velocity-based local variance could be calculated through $V_t^{\text{vel}} = \frac{1}{w} \sum_{i=-w/2}^{w/2} \|\mathbf{v}(t + i)\| $. Similarly, acceleration is the second derivative of trajectory position: $\mathbf{a}(t) = \frac{d^2\mathbf{p}(t)}{dt^2} $, which leads to the local motion variance estimation of $V_t^{\text{acc}} = \frac{1}{w} \sum_{i=-w/2}^{w/2} \|\mathbf{a}(t + i)\| $. With respect to the covariance-based local motion estimation, given a local window \( W_t = \{\mathbf{p}(t - w/2), \dots, \mathbf{p}(t + w/2)\} \), we compute: $
\boldsymbol{\Sigma}_t = \text{Cov}(W_t) $ and subsequently, the covariance-based motion variance is estimated as:
\begin{equation}
V_t^{\text{cov}} = \| \boldsymbol{\Sigma}_t \|_F
\end{equation}
where \( \| \cdot \|_F \) denotes the Frobenius norm.

When estimating the local motion variance through frequency features, we utilize the Short-Time Fourier Transform of pupil trajectory signal \( x(t) \) be $
Z_x(f, t) = \int x(\tau) w(\tau - t) e^{-j2\pi f \tau} d\tau $ of which the power spectrum is $
P_x(f, t) = |Z_x(f, t)|^2 $. Then, the frequency-domain motion variance is estimated as
$V_t^{\text{freq}} = \sqrt{ \text{Var}_f \left(P_x(f, t)\right) + \text{Var}_f \left(P_y(f, t)\right) }$.

In optical flow estimation as shown in Algorithm~\ref{algo:ofe}, we first estimate the appropriate size for the region of interest (ROI) around the filtered prediction using the first order derivatives of $x$ and $y$ and then, if the number of events within the selected ROI exceeds a set threshold, we accumulate and determine the cumulative vector trajectory of the events within ROI to softly shift the filtered prediction to further refine its spatial position. 

\setlength{\textfloatsep}{0pt}
\begin{algorithm}[t]
\caption{Motion-aware median filtering}\label{algo:median}
\begin{algorithmic}[1]
\REQUIRE Original predictions $\{x_{pred}, y_{pred}\}$, base window for local motion variance estimation $w_{base}$, minimum allowed smoothing window $w_{min}$, maximum allowed smoothing window $w_{max}$, percentile to determine adaptive window size $p$, method $f(.)$ $\in$ \{displacement, velocity, acceleration, covariance, frequency\}
\STATE Output: filtered predictions $\{x_{(f, pred)}, y_{(f, pred)}\}$ 
\STATE local motion variance $\longleftarrow$ $f(\{x_{pred}, y_{pred}\}, w_{base})$
\STATE smoothened variance $\longleftarrow$ rolling mean($w_{base}$, local motion variance)
\STATE median window $\longleftarrow$ clipping(smoothened variance, $w_{min}$, $w_{max}$)
\STATE adaptive windows $\longleftarrow$ clipping($w_{min}$, $w_{max}$, rolling(median window, $w_{base}$, $p$))
\STATE $\{x_{(f, pred)}, y_{(f, pred)}\}$ $\longleftarrow$ rolling median($\{x_{pred}, y_{pred}\}$, adaptive windows) 
\end{algorithmic}
\end{algorithm}

\begin{algorithm}[h!]
\caption{Rule-based optical flow estimation for smooth shifts}\label{algo:ofe}
\begin{algorithmic}[1]
\REQUIRE Continuous event stream with $N$ number of events $E^v_{i, (t,x,y,p)}$ where $i \in \{1, N\}$, filtered predictions $\{x_{(f, pred)}, y_{(f, pred)}\}$, scaling parameter $\tau$, count threshold $c$, difference threshold $\gamma$
\STATE Output: Refined predictions $\{x_{(R, f, pred)}, y_{(R, f, pred)}\}$
\STATE timestep $\longleftarrow$ $\frac{E^v(i=N, t_{max}) - E^v(i=1, t_{min})}{|\{x_{(f, pred)}, y_{(f, pred)}\}|}$
\STATE previous timestamp $\longleftarrow$ $E^v(i=1, t_{min}) \in E^v_{i, (t,x,y,p)}$
\STATE ROI size $R$ $\longleftarrow$ $\tau \times 10$
\FOR{$j$, $(x^j_{(f, pred)}, y^j_{(f, pred)})$ $\in$ $\{x_{(f, pred)}, y_{(f, pred)}\}$}
    \STATE current timestamp $\longleftarrow$ previous timestamp $+$ $(j+1) \times timestep$
    \IF{$j > c$}
        \STATE difference in x $\longleftarrow$ absolute($x^j_{(f, pred)} - mean(\{x^{j-c:j}_{(f, pred)}\})$)
        \STATE difference in y $\longleftarrow$ absolute($y^j_{(f, pred)} - mean(\{y^{j-c:j}_{(f, pred)}\})$)
        \IF{difference in x $> \tau \times \gamma$ $\cup$ difference in y $> \tau \times \gamma$}
            \STATE $R$ $\longleftarrow$ $(1+c) \times \tau$
        \ELSE
            \STATE $R$ $\longleftarrow$ $(1-c) \times \tau$
        \ENDIF
    \ENDIF
    \STATE events in ROI $\longleftarrow$ $E^v(t\in$ \{previous timestamp, current timestamp\}, $x \in \{x^j_{(f, pred)}-R, x^j_{(f, pred)}+R\}$, $y \in \{y^j_{(f, pred)}-R, y^j_{(f, pred)}+R\}$ $) \in E^v_{i, (t,x,y,p)}$
    \STATE previous timestamp $\longleftarrow$ current timestamp
    \STATE $n$ $\longleftarrow$ |events in ROI|
    \IF{$ n > \tau \times 10$}
        \STATE $dx \longleftarrow 0; dy \longleftarrow 0$
        \FOR{$k \in \{1, n\}$}
            \STATE $dx +=$ events in ROI($x=k$) $-$ events in ROI($x=k-1$)
            \STATE $dy +=$ events in ROI($y=k$) $-$ events in ROI($y=k-1$)
            \IF{absolute($dx$) $> 0 \cup$ absolute($dy$) $> 0$}
                \STATE $x^j_{(R, f, pred)} \longleftarrow x^j_{(f, pred)} + \frac{dx}{\Vert dx, dy\Vert}$
                \STATE $y^j_{(R, f, pred)} \longleftarrow y^j_{(f, pred)} + \frac{dx}{\Vert dx, dy\Vert}$
            \ENDIF
        \ENDFOR
    \ENDIF
\ENDFOR
\end{algorithmic}
\end{algorithm}
\section{Jitter Metric}
\label{sec: jitter_metric}

\subsection{Background \& Definition}
\label{ssec: jitter_definition}

While the existing metrics for evaluating pupil tracking performance are useful for assessing positional accuracy, they fall short in capturing the smooth continuity of the predicted pupil movements. This continuity is critical for the stable performance in downstream applications such as foveated rendering or gaze-based interaction~\cite{majaranta2014eye, sen2024eyetraes}, especially given the bounded and continuous nature of oculomotor (i.e., pupil movement) activity. Therefore, as a complementary metric, in this paper, we propose the following jitter metric (Eq.~\ref{eq: jitter}) to specifically evaluate the temporal smoothness of the predictions while also considering the true targets.

While designing the jitter metric, we ground our approach in two key premises that leverage pupil velocity on both global and local scales: (1) if the predicted trajectory differs substantially from the ground truth in how smoothly it evolves, their overall velocity distributions will show statistically significant differences, indicating that the predicted trajectory does not realistically reflect natural eye movement, and (2) sudden changes and irregular fluctuations in the predicted trajectory will be evident in its frequency spectrum, revealing fine-grained temporal mismatches compared to the true trajectory.

To operationalise (1) in our jitter metric, we use the Kullback-Leibler (KL) divergence to quantify information loss between the predicted and ground-truth velocity distributions. This captures global differences in motion realism: if the predicted velocity distribution is overly erratic or unnaturally smooth relative to the true distribution, the KL divergence, interpreted as a log-normalized comparative velocity entropy, will yield a higher jitter score.

To operationalise premise (2), we incorporate a spectral entropy (SPE) measure guided by spectral arc length (SPARC), following the framework of~\cite{balasubramanian2015analysis}. SPE is a frequency-domain metric that is less sensitive to high-frequency noise compared to traditional time-domain smoothness metrics (e.g., jerk). In this formulation, trajectories with more high-frequency content tend to exhibit longer arc lengths in the frequency domain, reflecting greater local irregularity. In contrast, smoother trajectories concentrate energy at lower frequencies, leading to simpler spectral profiles. To capture these local differences, we compute the spectral entropy of the predicted and ground-truth velocity signals and use their difference as a measure of local temporal deviation. The detailed derivation is presented in Section~\ref{sec: sparc_deviation}.
   
\begin{equation}
    JM(\text{pred}_{(x,y)}, \text{true}_{(x,y)}) = \lambda \cdot \frac{|\text{SPE}_{\frac{d(\text{pred})}{dt}} - \text{SPE}_{\frac{d(\text{true})}{dt}}|}{|\text{SPE}_{\frac{d(\text{true})}{dt}}| + \varepsilon} + 
(1-\lambda) \cdot \log\left(1 + D_{\text{KL}}\left(P_{f[\frac{d(\text{pred})}{dt}]} \,||\, P_{f[\frac{d(\text{true})}{dt}]}\right)\right)
    \label{eq: jitter}
\end{equation}

where, 
\begin{equation}
D_{\text{KL}}\left(P_{f[\frac{d(\text{pred})}{dt}]} \,||\, P_{f[\frac{d(\text{true})}{dt}]}\right) = \sum_{i} P_{f[\frac{d(\text{pred})}{dt}]}(i) \log \left( \frac{P_{f[\frac{d(\text{pred})}{dt}]}(i)}{P_{f[\frac{d(\text{true})}{dt}]}(i)} \right)
\label{eq: KL_div}
\end{equation}

\begin{equation}
\mathrm{SPE}\left(\frac{dg(x,y)}{dt}\right) = - \sum_{f > 0} \log(f + \varepsilon) \cdot \left( \frac{|V_f|}{\sum_f |V_f| + \varepsilon} \right)
\label{eq: sparc}
\end{equation}

Here, $\text{pred}_{(x,y)},\text{true}_{(x,y)}$ are predicted and true pupil trajectories whereas $f[.]$ is the function for velocity histogram estimation from predicted and true trajectories. The input to SPE as in Eq.~\ref{eq: sparc} is $g(x,y) \in \{\text{pred}_{(x,y)},\text{true}_{(x,y)}\}$ and $V_f$ is the Fourier magnitude of the respective velocity signal. The weight hyperparameter for balancing the impact of KL divergence and SPE terms is $\lambda \in [0, 1]$ and $\varepsilon$ is a small constant to ensure numerical stability (i.e., avoid division by zero). In summary, our jitter metric $JM$ is computed as a weighted sum of the normalized SPE difference and the log-normalized KL divergence of velocity histograms, and in addition, as designed, a lower value would reflect a more similar comparative temporal smoothness between the true and predicted pupil trajectories. Further, a detailed set of theoretical analysis on the proposed jitter metric, including the boundedness, continuity, formal constraints, continuity, and differentiability, is included in section~\ref{sec: theoretical}. 





\subsection{Derivation of SPE Equation}
\label{sec: sparc_deviation}

\subsubsection*{Original Discrete SPARC Definition}

Let $M(f)$ denote the normalized magnitude spectrum of the velocity signal $v(t)$, sampled at frequency $f_s$. Let $\{f_i, M_i\}$ be the discrete points of this spectrum such that $f_i$ being the frequency bin and $M_i$ being the normalized magnitude of Fourier transform at frequency $f_i$ (The velocity signal can be optionally filtered prior by a cutoff frequency $f_c$ and amplitude threshold $\theta$). $N$ is the number of frequency bins used. The SPARC metric is given by:

\begin{equation}
\text{SPARC} = -\sum_{i=1}^{N-1} \sqrt{
\left( \frac{f_{i+1} - f_i}{f_N - f_1} \right)^2 + 
\left( M_{i+1} - M_i \right)^2
}
\end{equation}

This corresponds to the total normalized negative arc length in the frequency-magnitude plane.

\subsubsection*{Step 1: Uniform Frequency Spacing Assumption}

Assuming the frequency bins are uniformly spaced, we define:

\[
\Delta f = f_{i+1} - f_i \text{ (is constant)}, \quad \Delta f' = \frac{\Delta f}{f_N - f_1}
\]

We then simplify the arc length as with $\Delta M_i = M_{i+1} - M_i$ and given that the spectrum is smooth $\Delta M_i < \delta$ where $\delta\,(> 0)$ is infinitesimally small and $\in \mathbb{R}$:

\begin{align}
\text{SPARC} &\approx -\Delta f' \sum_{i=1}^{N-1} \sqrt{
1 + \left( \frac{\Delta M_i}{\Delta f'} \right)^2 } \\
&= -\Delta f' \sum_{i=1}^{N-1} \left(
1 + \frac{1}{2} \left( \frac{\Delta M_i}{\Delta f'} \right)^2 \right)
\quad \text{(using } \sqrt{1 + x^2} \approx 1 + \frac{x^2}{2} \text{ when } |x| << 1 \text{)}\\
&= -(N - 1)\Delta f' - \frac{1}{2\Delta f'} \sum_{i=1}^{N-1} (\Delta M_i)^2
\end{align}

This shows that SPARC is negatively correlated with squared variation in spectral magnitude and thus, penalizes high spectral variation, which corresponds to non-smooth or jerky movements (Observation A).

\subsubsection*{Step 2: Frequency-Magnitude Reinterpretation}

We now reinterpret $M(f)$ as forming a normalized (discrete) probability distribution over frequency (instead of normalizing using the maximum amplitude), penalizing the higher spectral variation, such that $M(f) \approx |V_f|$:

\begin{equation}
P(f_i) = \frac{M(f_i)}{\sum_{j} M(f_j)}
\end{equation}

Motivated by Observation A, we propose a frequency-weighted sum form that emphasizes smoothness by penalizing energy concentrated at high frequencies (Note that the frequency weighting term $log(f)$ penalizes the higher frequencies by leading the SPE towards more negative sum):

\begin{equation}
\text{SPARC} \approx \text{SPE} = - \sum_{i} \log(f_i + \epsilon) \cdot P(f_i)
\end{equation}

where $\epsilon > 0$ is a small constant added for numerical stability.

Even though this version drops the explicit arc-length geometry, it retains the original spirit of SPARC by favoring low-frequency spectral energy concentration, and is more computationally tractable and differentiable. 

\subsubsection*{Summary}

We conclude with the following approximate proxy metric:

\begin{equation}
\text{SPE} = - \sum_{i} \log(f_i + \epsilon) \cdot \frac{M(f_i)}{\sum_j M(f_j)}
\end{equation}

\subsection{Theoretical Analysis on Jitter Metric}
\label{sec: theoretical}

\subsubsection{Theoretical Justification}
Supplementary to our explanations in section~\ref{ssec: jitter_definition} and the derivation in section~\ref{sec: sparc_deviation}, below we present two lemmas to show why the jitter metric captures the smoothness.

\begin{lemma}
    Let $f(t)$ be a continuous-time signal. Then, the smoother $f(t)$ is, the more concentrated its frequency spectrum is in low frequencies. As a result, the spectral entropy is lower.
\end{lemma}

\begin{proof}
    The underlying intuition behind the lemma is that smoother signals exhibit fewer rapid fluctuations, which corresponds—in the Fourier domain—to reduced magnitude in high-frequency components. The spectral entropy (SPE), used in the first term of our jitter metric, quantifies the dispersion of energy across frequency bands. A highly smooth signal concentrates energy in lower frequencies and thus yields a lower SPE, reflecting this property in an absolute sense.

    Formally, in more general sense, if the velocity function $f \in C^k(\mathbb{R})$ with $k-$continuous derivatives: $f, tf(t) \in L^1(\mathbb{R})$ and Fourier transform of $f$, $\hat{f} \in C^1(\mathbb{R})$, then by differentiation theorem, 
    \begin{equation}
        \hat{tf(\gamma)} = \frac{-1}{2\pi i}\frac{d\hat{f}}{d\gamma}(\gamma)
    \end{equation}
    Similarly, if $t^kf(t) \in L^1(\mathbb{R})$ for some $k \in \mathbb{N}$, then, $\hat{f} \in C^k(\mathbb{R})$ for $0 \leq j \leq k$,
    \begin{equation}
        \hat{x^jf(\gamma)} = \left(\frac{-1}{2\pi i}\right)^j\frac{d^j\hat{f}}{d\gamma^j}(\gamma)
    \end{equation}
    
    This \textit{decay} property implies that smoother signals, characterized by larger values of $k$, exhibit faster spectral decay, with their frequency components approaching zero as frequency tends to infinity. 

    Moreover, the $\log$ penalizing term in $SPE$ in the first term in jitter metric amplifies the contribution of high-frequency components -- since $\log(f+\epsilon)$ is monotonically increasing for $f+\epsilon > 1$, energy concentrated at higher frequencies incurs a larger penalty. 

    Therefore, considering the above, a smoother signal $f(t)$ has a faster decaying spectrum and thus lower weights on high frequency terms, and thus, has lower spectral entropy.  
\end{proof}

\begin{lemma}
    Let $P$ and $Q$ be discrete velocity magnitude distributions for predicted and true trajectories. Then, $D_{KL}(P||Q)$ quantifies how different the temporal dynamics are between the predicted and true trajectories, with larger divergence implying greater discrepancy in movement similarity. 
\end{lemma}

\begin{proof}
    The core intuition behind the lemma is that if the predicted trajectory captures motion variability and regularity similar to the ground truth, then fewer additional bits are required to encode samples from distribution $P$ using a model based on $Q$.

    Formally, assuming both $P(i), Q(i) > 0$ and $\sum_{i} P(i) = \sum_{i} Q(i) = 1$, and noting that the KL divergence has the following properties (from Gibb's inequality), 
    \begin{equation}
        D_{KL}(P||Q) \geq 0\,\, \text{and}\,\, D_{KL}(P||Q)=0 \longleftrightarrow P=Q 
    \end{equation}
    In addition, as shown in theorem~\ref{theo: upper_bound}, KL divergence is finite only when the support of $P$ is contained within the support of $Q$. For instance, if $P$ assigns a probability to velocities where $Q$ is \textit{almost} $0$, then, $D_{KL}(P||Q) \longrightarrow \infty$. 

    Based on the above properties and the log-likelihood ratio, it implies that, 
    \begin{itemize}
        \item $P(i) >> Q(i)$ for some $i\in\chi \longrightarrow \frac{P(i)}{Q(i)} >> \delta$, which in turn implies that $D_{KL}$ increases. 
        \item Similarly, $P(i) << Q(i)$ for some $i\in\chi \longrightarrow \log{\frac{P(i)}{Q(i)}} << 0$ while $P(i) \approx 0$. 
    \end{itemize}
    As a case study, consider a  simple family of probability distributions parametrized by $\theta$. Let $P_{\theta = 0} = Q$ and therefore, $P_{\theta}$ differs from $Q$ as $\theta$ is modified. If $\theta = 0\,\,\text{and}\,\, \theta \in \mathbb{N}$ and the family of probability distributions be univariate Gaussians, then, $Q$ is characterized by $\mathcal{N}(0, \sigma)$ and generally, $P_\theta = \mathcal{N}(\theta, \sigma)$. Therefore, it is trivial that as $\theta$ grows, $P_\theta$ shifts away from $Q$. Then, from general case for $D_{KL}$ of univariate Gaussians,
    \begin{equation}
        D_{KL}(P||Q) = \log{\frac{\sigma_2}{\sigma_1}}+\frac{\sigma_1^2+(\mu_1-\mu_2)^2}{2\sigma_2^2}-\frac{1}{2}
    \end{equation}
    Since in our case, $\sigma_1=\sigma_2$, $\mu_1 = \theta$, and $\mu_2=0$, then, 
    \begin{equation}
        D_{KL}(P||Q) = \log\frac{\sigma_1}{\sigma_1} + \frac{\sigma_1^2+(\theta -0)^2}{2\sigma_1^2} - \frac{1}{2} = \frac{\theta^2}{2\sigma_1^2}
    \end{equation}
    Therefore, it is trivial that $D_{KL}$ increases monotonically with $\theta$, i.e., when $P_\theta$ differs more from $Q$. Similarly, a more generic proof can follow Donsker and Varadhan's variational formula in this context.  
\end{proof}

\subsubsection{Formal Constraints}
Jitter metric is theoretically plausible under the following set of assumptions:
\begin{itemize}
    \item The predicted and true trajectories must be at least $C^1-$continuous.
    \item The jitter metric must have the full support of $P_{f[\frac{d(\text{true})}{dt}]}$.
    \item Both predicted and true trajectories must have bounded energies: $\int |\hat{f}(.)|^2 d\omega < \infty$
    \item Additionally, as required by the task at hand, the velocity trajectories should be non-negative and \textit{temporally aligned}.  
\end{itemize}

\begin{lemma}
    Let $\text{pred}_{(x,y)} (t) \in C^1(\mathbb{R})$ be the predicted pupil trajectory, then, the velocity $\frac{d\text{pred}_{(x,y)}(t)}{dt}$ is continuous, and the velocity histogram $P_v$ is well-defined as a Radon-Nikodym derivative with respect to the Lebesgue measure. 
    \label{lemma: histogram}
\end{lemma}

\begin{proof}
    By definition of $C^1-$continuity, $\text{pred}_{(x,y)} (t)$ is differentiable (on a closed and bounded interval), and $\frac{d\text{pred}_{(x,y)}(t)}{dt}$ is continuous. Continuity of $\frac{d\text{pred}_{(x,y)}(t)}{dt}$ ensures it is measurable, by measure theory.

    Since $\text{pred}_{(x,y)} (t)$ is $C^1$, then, $\frac{d\text{pred}_{(x,y)}(t)}{dt}$ is bounded on any compact interval $[a,b]; a,b \in \mathbb{R}$ by extreme value theorem. Further, $\frac{d\text{pred}_{(x,y)}(t)}{dt}$ does not diverge since it has a finite energy, i.e., $\int |\frac{d\text{pred}_{(x,y)}(t)}{dt}|^2 dt < \infty$ (as it is a physical trajectory, i.e., a pupil velocity trajectory). 

    Considering the constructed velocity histogram $P_v$ on the time interval $[0,T]\,\,T\in\mathbb{R}$, $\mu$ is the Lebesgue measure, and $A\subseteq \mathbb{R}$ is any measurable set:
    \begin{equation}
        P_v(A) = \frac{1}{T}\mu(\{t\in[0,T]|\frac{d\text{pred}_{(x,y)}(t)}{dt} \in A\}) 
    \end{equation}
    As $\frac{d\text{pred}_{(x,y)}(t)}{dt}$ is continuous, then measurable, the preimage $\{t\in[0,T]|\frac{d\text{pred}_{(x,y)}(t)}{dt} \in A\}$ is Lebesgue-measurable for any Borel set $A$. Further, as $\frac{d\text{pred}_{(x,y)}(t)}{dt}$ is bounded and $T< \infty$ (due to being characteristics of physical trajectories), $P_v$ is a valid probability measure. 

    Being a physical trajectory, $\frac{d\text{pred}_{(x,y)}(t)}{dt}$ is piecewise monotonic. Therefore, it is possible to further extend that $P_v$ should admit a probability density function $p(v)$:
    \begin{equation}
        p(v) = \frac{1}{T} \sum_{t_i|\frac{d\text{pred}_{(x,y)}(t_i)}{dt}=v} |\frac{dv}{dt}(t_i)|^{-1}
    \end{equation}
    where the sum is over roots $t_i$ of $\frac{d\text{pred}_{(x,y)}(t)}{dt} - v = 0$. As a practical implication, velocity from discrete pupil trajectories (of $N$ samples) can be computed via finite differences: $v_i = \frac{\text{pred}_{(x,y)} (t_{i+1})-\text{pred}_{(x,y)} (t_i)}{t_{i+1}-t_i}$, as if $\text{pred}_{(x,y)} (t)$ is $C^1$, then, the histogram of $\{v_i\}$ converges to $P_v$ as $N\longrightarrow \infty$. Further, a similar proof can follow true pupil trajectories as well. 
\end{proof}

\subsubsection{Scale Invariance}

\begin{theorem}
    Jitter metric is scale-invariant for $\forall \alpha > 0$ under negligible $\varepsilon$. 
\end{theorem}

\begin{proof}
    Consider a linear and consistent scaling factor $\alpha\,(>0)$ across both predicted and true trajectories. Then, the scaled trajectories are: $\text{pred}'_{(x,y)} (t) = \alpha \cdot \text{pred}_{(x,y)} (t)$ and $\text{true}'_{(x,y)} (t)= \alpha \cdot \text{true}_{(x,y)} (t)$. Therefore, the velocities scale as: $\frac{d\text{pred}'_{(x,y)}(t)}{dt} = \alpha \cdot \frac{d\text{pred}_{(x,y)}(t)}{dt}$ and $\frac{d\text{true}'_{(x,y)}(t)}{dt} = \alpha\cdot\frac{d\text{true}_{(x,y)}(t)}{dt}$. By extension, the velocity distributions are: $P_{\text{pred}'}(v) = \frac{1}{\alpha}P_{\text{pred}}(\frac{v}{\alpha})$ and $P_{\text{true}'}(v) = \frac{1}{\alpha}P_{\text{true}}(\frac{v}{\alpha})$. 

    When considering the first term in the jitter metric, from the scaling property of Fourier transform: $\mathcal{F}( \frac{d\text{pred}'_{(x,y)}(t)}{dt}) = \mathcal{F}(\alpha \frac{d\text{pred}_{(x,y)}(t)}{dt}) = \alpha \mathcal{F}(\frac{d\text{pred}_{(x,y)}(t)}{dt}) \longrightarrow |V'_f| = \alpha |V_f|$. Therefore, 
    \begin{equation}
        \text{SPE}(\frac{d\text{pred}'_{(x,y)}(t)}{dt}) = - \sum_{f > 0} \log(f + \varepsilon) \cdot \left( \frac{|V'_f|}{\sum_f |V'_f| + \varepsilon} \right) = - \sum_{f > 0} \log(f + \varepsilon) \cdot \left( \frac{\alpha \cdot |V_f|}{\sum_f \alpha \cdot |V_f| + \varepsilon} \right)
    \end{equation}
    \begin{equation}
        = - \sum_{f > 0} \log(f + \varepsilon) \cdot \left( \frac{\alpha \cdot |V_f|}{\alpha \sum_f |V_f| + \varepsilon} \right) \approx - \sum_{f > 0} \log(f + \varepsilon) \cdot \left( \frac{\alpha \cdot |V_f|}{\alpha\sum_f |V_f|} \right)\,\,\text{under negligible}\,\,\varepsilon
    \end{equation}
    \begin{equation}
        \approx - \sum_{f > 0} \log(f + \varepsilon) \cdot \left( \frac{ \cdot |V_f|}{\sum_f |V_f| + \varepsilon} \right) = \text{SPE}(\frac{d\text{pred}_{(x,y)}(t)}{dt})
    \end{equation}
    The same can be proved for true trajectory as well. Therefore, the first term is scale-invariant for $\alpha>0$ for negligible $\varepsilon$.

    Regarding the second term, for scaled $P'$ and $Q'$, while adapting the continuous definition for KL divergence, 
    \begin{equation}
        D_{KL}(P'||Q'') = \int P'(v)\log\frac{P'(v)}{Q'(v)}dv = \int \frac{1}{\alpha} P(\frac{v}{\alpha})\log\frac{\frac{1}{\alpha}P(\frac{v}{\alpha})}{\frac{1}{\alpha}Q(\frac{v}{\alpha})} dv
    \end{equation}
    Changing variables: $u = \frac{v}{\alpha}$:
    \begin{equation}
        = \int \frac{1}{\alpha} P(u)\log\frac{P(u)}{Q(u)}\alpha\cdot du = \int P(u)\log\frac{P(u)}{Q(u)}du = D_{KL}(P||Q)
    \end{equation}
    Therefore, $D_{KL}$ is scale-invariant, so is $\log(1+D_{KL})$. 

    Therefore, from all above, the jitter metric is scale invariant for $\forall\alpha>0$ with negligible $\varepsilon$.
\end{proof}

\subsubsection{Lower Bound}
\begin{theorem}
Jitter metric is lower bounded. In other words, for any predicted ($\text{pred}_{(x,y)}$) and true ($\text{true}_{(x,y)}$) trajectories, the metric satisfies $JM(\text{pred}_{(x,y)}, \text{true}_{(x,y)}) \geq 0$ with equality iff $\text{pred}_{(x,y)} = \text{true}_{(x,y)}$
\end{theorem}

\begin{proof}
Both first and second terms in the jitter metric are strictly \textit{non-negative} due to both additive terms being non-negative (while $\lambda \in [0, 1] \text{ and } \lambda \in \mathbb{R}$):
\begin{itemize}
    \item The first term is trivially non-negative (i.e., $|\text{SPE}_{\frac{d(\text{pred})}{dt}} - \text{SPE}_{\frac{d(\text{true})}{dt}}| \geq 0 $)
    \item The second term is also non-negative since KL divergence is always non-negative (i.e., Gibb's inequality, $D_{KL}(P||Q) \geq 0$ leads to $\log(1+x) \geq 0 \text{ for } x \geq 0$). 
\end{itemize} 

Regarding the condition for equality,
\begin{itemize}
    \item When the predicted trajectory and the true trajectory are congruent in 1D sense (i.e., $\text{pred}_{(x,y)} = \text{true}_{(x,y)}$), then, $\frac{d(\text{pred})}{dt} = \frac{d(\text{true})}{dt}$. Therefore, $|\text{SPE}_{\frac{d(\text{pred})}{dt}} - \text{SPE}_{\frac{d(\text{true})}{dt}}| = 0$ and the first term in jitter metric vanishes. 
    \item Similarly, $\frac{d(\text{pred})}{dt} = \frac{d(\text{true})}{dt}$ leads to $D_{KL}(.) = 0$, since $\log \left( \frac{P_{f[\frac{d(\text{pred})}{dt}]}(i)}{P_{f[\frac{d(\text{true})}{dt}]}(i)} \right) = 0$, and therefore, the second term also vanishes.
\end{itemize}

Since both first and seconds terms are \textit{strictly} non-negative, their weighted sum (i.e., jitter metric value) is also \textit{strictly} non-negative. Further, $JM(.) = 0$ is only possible when $\text{pred}_{(x,y)} = \text{true}_{(x,y)}$ which is the lower bound for the metric. 
\end{proof}

\subsubsection{Upper Bound}
\label{sssec: upper_bound}

\begin{theorem}
    Jitter metric is \textit{not} upper-bounded.
    \label{theo: upper_bound}
\end{theorem}

\begin{proof}
    The first term in the jitter metric is bounded if both $\text{SPE}_{\frac{d(\text{true})}{dt}}$ and $\text{SPE}_{\frac{d(\text{pred})}{dt}}$ are finite and $\in \mathbb{R}$. 

    Regarding the second term: let $P_{f[\frac{d(\text{pred})}{dt}]}$ be a velocity distribution with support disjoint from $P_{f[\frac{d(\text{true})}{dt}]}$. In other words, $P_{f[\frac{d(\text{pred})}{dt}]}$ assigns non-zero probability to events where $P_{f[\frac{d(\text{true})}{dt}]}$ has zero probability. Then, $D_{KL}(.) \longrightarrow \infty$, so, $JM(.) \geq (1-\lambda) \cdot \log(1 + \infty) = \infty$.

    To ensure the sufficient condition for KL divergence to be upper-bounded, 
    \begin{lemma}
        Given both $P_{f[\frac{d(\text{true})}{dt}]}$ and $P_{f[\frac{d(\text{pred})}{dt}]}$ have the same support $\chi$ and $P_{f[\frac{d(\text{pred})}{dt}]}$ has a finite upper bound, then $D_{KL}(P_{f[\frac{d(\text{pred})}{dt}]}||P_{f[\frac{d(\text{true})}{dt}]}) < \infty$
    \end{lemma}

    \begin{proof}
        Since $P_{f[\frac{d(\text{true})}{dt}]}$ has compact support:
        \begin{equation}
            \underline{q} = \inf_{i\in \chi} P_{f[\frac{d(\text{true})}{dt}]}(i) > 0
        \end{equation}
        Similarly, since $P_{f[\frac{d(\text{pred})}{dt}]}$ has compact support, 
        \begin{equation}
            \bar{p} = \sup_{i\in \chi} P_{f[\frac{d(\text{pred})}{dt}]}(i) > 0
        \end{equation}
        Since both are on the same support and $P_{f[\frac{d(\text{pred})}{dt}]}$ is bounded, 
        \begin{equation}
            0 < \underline{q} \leq \bar{p} < \infty
        \end{equation}
        \begin{equation}
            \sup_{i\in\chi} \log{\left(\frac{P_{f[\frac{d(\text{pred})}{dt}]}(i)}{P_{f[\frac{d(\text{true})}{dt}]}(i)}\right)} \leq \log{\bar{p}}-\log{\underline{q}}
        \end{equation}
        Let $L = \log{\bar{p}}-\log{\underline{q}}$, then, 
        \begin{equation}
            D_{KL}(P_{f[\frac{d(\text{pred})}{dt}]}||P_{f[\frac{d(\text{true})}{dt}]}) = \sum_{i\in \chi} P_{f[\frac{d(\text{pred})}{dt}]}(i)\log\left(\frac{P_{f[\frac{d(\text{pred})}{dt}]}}{P_{f[\frac{d(\text{true})}{dt}]}}\right) \leq \sup_{i\in\chi} \log{\left(\frac{P_{f[\frac{d(\text{pred})}{dt}]}(i)}{P_{f[\frac{d(\text{true})}{dt}]}(i)}\right)}\sum_{i\in\chi}P_{f[\frac{d(\text{pred})}{dt}]}(i)
        \end{equation}
        \begin{equation}
            \leq (\log{\bar{p}}-\log{\underline{q}})\sum_{i\in\chi}P_{f[\frac{d(\text{pred})}{dt}]}(i) = L < \infty 
        \end{equation}
    \end{proof}

    Therefore, if $P_{f[\frac{d(\text{pred})}{dt}]}$ is with support disjoint from $P_{f[\frac{d(\text{true})}{dt}]}$, then, second term dominates and the metric becomes unbounded above. 
\end{proof}

\subsubsection{Continuity}

\begin{theorem}
    Jitter metric is continuous everywhere except when $P_{f[\frac{d(\text{true})}{dt}]}$ has zero-mass bins (i.e., if not smoothened). In other words, with full support, jitter metric is continuous. 
    \label{theo: continuity}
\end{theorem}

\begin{proof}
    Assume both $\text{pred}_{(x,y)}$ and $\text{true}_{(x,y)}$ are differentiable on a closed and bounded interval (Note that these are valid assumptions given the typical ocularmotor i.e., pupil activity, function~\cite{bandara2024eyegraph}). Then, both $\frac{d(\text{pred})}{dt}$ and $\frac{d(\text{true})}{dt}$ are continuous functions under $L^2$ norm: if $\Vert \text{true} - \text{pred} \Vert 
 \longrightarrow 0 \text{ then } \Vert \frac{d(\text{true})}{dt} - \frac{d(\text{pred})}{dt} \Vert \longrightarrow 0$.

    Based on the assumptions of differentiability on a closed and bounded interval (and thereby continuous since differentiability implies continuity), both $\frac{d(\text{pred})}{dt}$ and $\frac{d(\text{true})}{dt}$ are Riemann integrable, which in tern implies that those are Lebesgue integrable. Therefore, both velocity trajectories are in $L^1$ space: $\frac{d(\text{pred})}{dt} \in L^1 (\mathbb{R}^n)$ and $\frac{d(\text{true})}{dt} \in L^1 (\mathbb{R}^n)$.

    If $\frac{d(\text{pred}_{(x)})}{dt} = f_{pred(x)}(t)$, let Fourier transform of predicted velocity trajectory be: \begin{equation}
 \hat{f}_{pred(x)}(\gamma) = \int_{\mathbb{R}^n} e^{-i\gamma z}f_{pred(x)}(z) dz\,;\,\, \gamma \in \mathbb{R}
    \end{equation}. Note that, for simplicity, we only consider 1D case (for $x$ coordinate) considering one variable at a time in the trajectory and this is easily extendable to multi-variable case as the Fourier transform on multivariables is well-defined. $\forall \gamma_n \longrightarrow \gamma$:
    \begin{equation}
        g_n(z) = e^{-i\gamma_nz}f_{pred(x)}(z)\,\,\text{is measurable}\,\,\forall n\in \mathbb{N}
    \end{equation}
    \begin{equation}
        g_n(z) \longrightarrow e^{-i\gamma z}f_{pred(x)}(z) = g(z)\,\,\forall z
    \end{equation}
    \begin{equation}
        |e^{-i\gamma_nz}f_{pred(x)}(z)|= |f_{pred(x)}(z)|\,\,\text{where}\,\,|f_{pred(x)}(z)|\,\,\text{is integrable}
    \end{equation}
Therefore, by dominated convergence theorem, 
\begin{equation}
    \lim_{n \longrightarrow\infty} \hat{f}_{pred(x)}(\gamma_n) = \lim_{n \longrightarrow\infty} \int_{\mathbb{R}^n} e^{-i\gamma_nz}f_{pred(x)}(z) dz = \lim_{n \longrightarrow\infty} \int_{\mathbb{R}^n} g_n(z) dz = \int_{\mathbb{R}^n}g(z) dz
\end{equation}
\begin{equation}
   \lim_{n \longrightarrow\infty} \hat{f}_{pred(x)}(\gamma_n) = \int_{\mathbb{R}^n}e^{-i\gamma z}f_{pred(x)}(z)dz = \hat{f}_{pred(x)}(\gamma)
\end{equation}
Therefore, $\hat{f}_{pred(x)}(\gamma)$ is \textit{uniformly} continuous. Similarly, we can prove this for true trajectory as well under the same set of assumptions. Since taking the absolute value, normalization, multiplication, and $\log$ (for $x>0$) do not violate the continuity, the first term is continuous.

Assuming soft histogramming (i.e., soft binning or kernel density estimation), in other words, if $P_{f[\frac{d(\text{true})}{dt}]}$ has full support: $P_{f[\frac{d(\text{true})}{dt}]} > 0$ everywhere, then, 
\begin{equation}
P_{f[\frac{d(\text{pred})}{dt}]}, P_{f[\frac{d(\text{true})}{dt}]} \in (0, 1] \longrightarrow P_{f[\frac{d(\text{pred})}{dt}]}/ P_{f[\frac{d(\text{true})}{dt}]}\,\,\text{is continuous}
\end{equation}
As $\log$ is continuous over $(0, \infty)$, by extension, under the said assumption, the second term is also continuous. 

As the convex sum of continuous functions is also continuous, the jitter metric is continuous while with full support.  
\end{proof}

\subsubsection{Differentiability}
\begin{theorem}
    Jitter metric is differentiable almost everywhere except when (1) $SPE_{\text{pred}} = SPE_{\text{true}}$, (2) $V_f = 0$ for any $f$, (3) with support disjoint from $P_{f[\frac{d(\text{true})}{dt}]}$, and (4) $P_{f[\frac{d(\text{pred})}{dt}]}$ is on the simplex boundary. 
\end{theorem}

\begin{proof}
    Assume both $\text{pred}_{(x,y)}$ and $\text{true}_{(x,y)}$ are differentiable on a closed and bounded interval (Note that these are valid assumptions given the typical ocularmotor i.e., pupil activity, function~\cite{bandara2024eyegraph} as a physical signal). Then, both $\frac{d(\text{pred})}{dt}$ and $\frac{d(\text{true})}{dt}$ are continuous functions under $L^2$ norm: if $\Vert \text{true} - \text{pred} \Vert 
 \longrightarrow 0 \text{ then } \Vert \frac{d(\text{true})}{dt} - \frac{d(\text{pred})}{dt} \Vert \longrightarrow 0$.

    Based on the assumptions of differentiability on a closed and bounded interval (and thereby continuous since differentiability implies continuity), both $\frac{d(\text{pred})}{dt}$ and $\frac{d(\text{true})}{dt}$ are Riemann integrable, which in tern implies that those are Lebesgue integrable. Therefore, both velocity trajectories are in $L^1$ space: $\frac{d(\text{pred})}{dt} \in L^1 (\mathbb{R}^n)$ and $\frac{d(\text{true})}{dt} \in L^1 (\mathbb{R}^n)$.

    If $\frac{d(\text{pred}_{(x)})}{dt} = f_{pred(x)}(t)$, let Fourier transform of predicted velocity trajectory be: \begin{equation}
 \hat{f}_{pred(x)}(\gamma) = \int_{\mathbb{R}^n} e^{-i\gamma z}f_{pred(x)}(z) dz\,;\,\, \gamma\in \mathbb{R}
    \end{equation}. Note that, for simplicity, we only consider 1D case (for $x$ coordinate) considering one variable at a time in the trajectory and this is easily extendable to multi-variable case as the Fourier transform on multivariables is well-defined. 

    Following the theorem~\ref{theo: continuity} on the continuity of the jitter metric, more specifically, $\hat{f}_{pred(x)}(\gamma)$ is \textit{uniformly} continuous, and assuming $tf_{pred(x)}(t) \in L^1(\mathbb{R})$, define the function $g_{pred(x)}(t)$ as $\hat{-it f_{pred(x)}(t)}$ which is also \textit{uniformly} continuous. Consider, 
    \begin{equation}
        \int_{0}^{y} g_{pred(x)}(h) dh = \int_0^y \int_{\mathbb{R}} -ite^{-ith}f_{pred(x)}(t)\,dt\,dh
        = \int_{\mathbb{R}} e^{-ith}|_0^yf_{pred(x)}(t)dt 
    \end{equation}
    \begin{equation}
        = \int_{\mathbb{R}}f_{pred(x)}(t)[e^{-ity}-e^{it\cdot0}]dt = \hat{f}_{pred(x)}(y) -\hat{f}_{pred(x)}(0)
    \end{equation}
    Therefore, by fundamental theorem of calculus, $\hat{f}_{pred(x)}(\gamma)$ is differentiable almost everywhere. Similarly, we can prove this result for $\hat{f}_{pred(y)}(\gamma)$, and the true velocity trajectory as well. 

    Upon the immediate above result, the first term in the jitter metric is differentiable almost everywhere except when $|V_f|=0$ since the absolute value function $\in \mathbb{R}$ is not differentiable at $0$. Further, since $SPE_{\text{pred}}(.) = SPE_{\text{true}}$ leads to vanishing the first term, the differentiability of jitter metric is not defined when $SPE_{\text{pred}}(.) = SPE_{\text{true}}$ as well. 

    Regarding the second term in the jitter metric, if $P_{f[\frac{d(\text{pred})}{dt}]}$ and $P_{f[\frac{d(\text{true})}{dt}]}$ can be parametrized using parameters $\theta \in \mathbb{R}^n$ and $\psi \in \mathbb{R}^m$ respectively, then, $P_{f[\frac{d(\text{pred})}{dt}]}$ can be written as $P_{f[\frac{d(\text{pred})}{dt}](x;y|\theta)}$, whereas $P_{f[\frac{d(\text{true})}{dt}]}$ as $P_{f[\frac{d(\text{true})}{dt}](x;y|\psi)}$. Here, we assume that $P_{f[\frac{d(\text{true})}{dt}](x;y|\theta)}, P_{f[\frac{d(\text{true})}{dt}](x;y|\psi)} \in C^1(\mathbb{R})$ on the space $\chi$ and $P_{f[\frac{d(\text{true})}{dt}](x;y|\psi)} > 0\,\,\forall(x;y)\in\chi\,\,\text{and}\,\,\psi$. For simplicity, we denote $P_{f[\frac{d(\text{true})}{dt}](x;y|\psi)}$ and $P_{f[\frac{d(\text{true})}{dt}](x;y|\psi)}$ as $P_{\psi}$ and $P_\theta$ respectively. Therefore,
    \begin{equation}
        D_{KL}(P_\theta||P_\psi) = \sum_{i\in\chi}P_\theta\log(\frac{P_\theta}{P_\psi})
    \end{equation}
    For a fixed $P_{\psi}$ wrt $\theta_l$, 
    \begin{equation}
        \frac{\partial}{\partial \theta_l} D_{KL}(P_\theta||P_\psi) = \sum_{i \in \chi} \left[ \frac{\partial P_\theta (i)}{\partial \theta_l}\log \left( \frac{P_\theta (i)}{P_\psi (i)}\right) +P_\theta (i) \frac{\partial}{\partial \theta_l} \log \left( \frac{P_\theta (i)}{P_\psi (i)}\right) \right]
    \end{equation}
    \begin{equation}
        = \sum_{i \in \chi} \left[ \frac{\partial P_\theta (i)}{\partial \theta_l}\log \left( \frac{P_\theta (i)}{P_\psi (i)}\right) + P_\theta (i) \left\{ \frac{\partial}{\partial \theta_l } (\log P_\theta (i) - \log P_\psi (i)) \right\}\right]
    \end{equation}
    \begin{equation}
        = \sum_{i \in \chi} \left[ \frac{\partial P_\theta (i)}{\partial \theta_l}\log \left( \frac{P_\theta (i)}{P_\psi (i)}\right) + P_\theta (i) \frac{1}{P_\theta (i)}\frac{\partial P_\theta (i)}{\partial \theta_l} \right]\,\,\text{i.e., a fixed $P_{\psi}$ wrt $\theta_l$}
    \end{equation}
    \begin{equation}
       \frac{\partial}{\partial \theta_l} D_{KL}(P_\theta||P_\psi) = \sum_{i \in \chi} \left[ \frac{\partial P_\theta (i)}{\partial \theta_l} \left(\log \left( \frac{P_\theta (i)}{P_\psi (i)}\right) +1 \right) \right]
    \end{equation}
\end{proof}
Therefore, under the assumptions of $P_\theta, P_\psi \in C^1$ in $\theta, \psi$, respectively, on $\chi$ and $P_\psi (i) > 0\,\,\forall i\in\chi$ (and given $P_\theta$ is not on the simplex boundary), the above sum uniformly converges and the derivative exists and continuous. Similarly, the partial derivative of $D_{KL}$ wrt $\psi_m$ can be proven to be:
\begin{equation}
    \frac{\partial}{\partial \psi_m} D_{KL}(P_\theta||P_\psi) = -\sum_{i\in\chi}P_\theta (i)\frac{\partial}{\partial \psi_m} \log P_\psi (i)
\end{equation}
which exists and continuous under the same set of assumptions. 

A more detailed proof with gradient derivations can follow as proved in variational bayes~\cite{kingma2022autoencodingvariationalbayes}.  



\subsubsection{Time Complexity}
The dominant operation in the first term is the Fourier transform. If used fast Fourier transform (FFT) on an array of length $n$, then,
\begin{equation}
    \text{Time complexity of FFT} = O(n\log n)
\end{equation}
As taking magnitude, normalization, weighted sum are in $O(n)$, then, the time complexity of the first term:
\begin{equation}
    \text{Time complexity of the first term} = O(n\log n + n) = O(n \log n)
\end{equation}

For the second term, assuming histograms with $b$ bins ($b << n$), soft-binning is of order $O(n\cdot b)$ and (discrete) KL divergence is of order $O(b)$. Subsequently, 
\begin{equation}
    \text{Time complexity of the second term} = O(n\cdot b + b) = O(n\cdot b)
\end{equation}
Therefore, 
\begin{equation}
    \text{Time complexity of Jitter Metric} = O(n\log n + n\cdot b) = O(n\log n)
\end{equation}




\subsection{Empirical Justification for Jitter Metric}
\label{ssec: jitter_metric}

\begin{figure}[!h]
    \centering
    \begin{subfigure}[t]{0.6\linewidth}
        \centering
        \includegraphics[width=\linewidth]{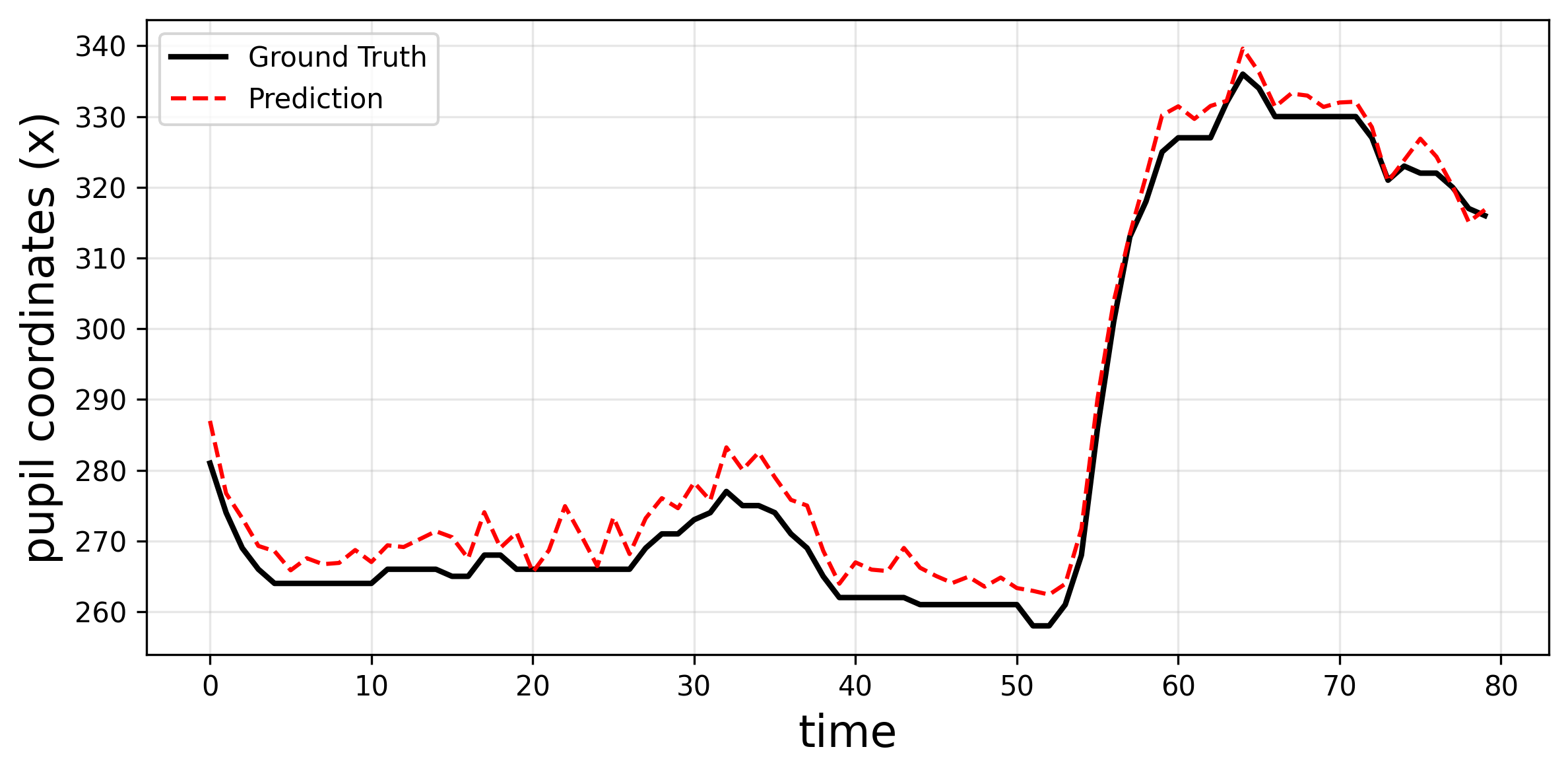}
        \caption{MSE: 15.83, JM: 0.18}
        \label{fig: (a)}
    \end{subfigure}
    \hfill
    \begin{subfigure}[t]{0.48\linewidth}
        \centering
        \includegraphics[width=\linewidth]{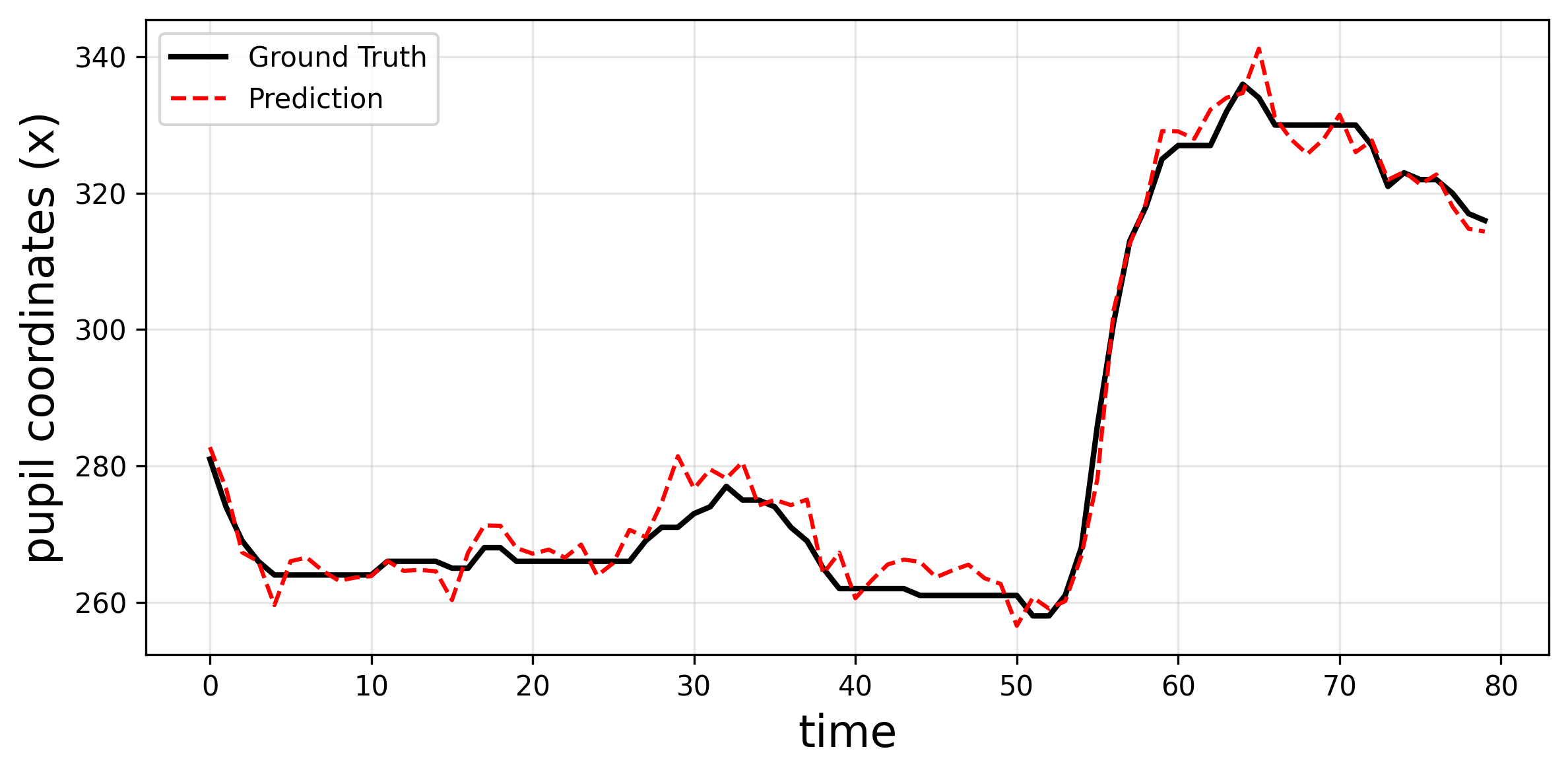}
        \caption{MSE: 9.82, JM: 0.23}
        \label{fig: (b)}
    \end{subfigure}
        \hfill
    \begin{subfigure}[t]{0.48\linewidth}
        \centering
        \includegraphics[width=\linewidth]{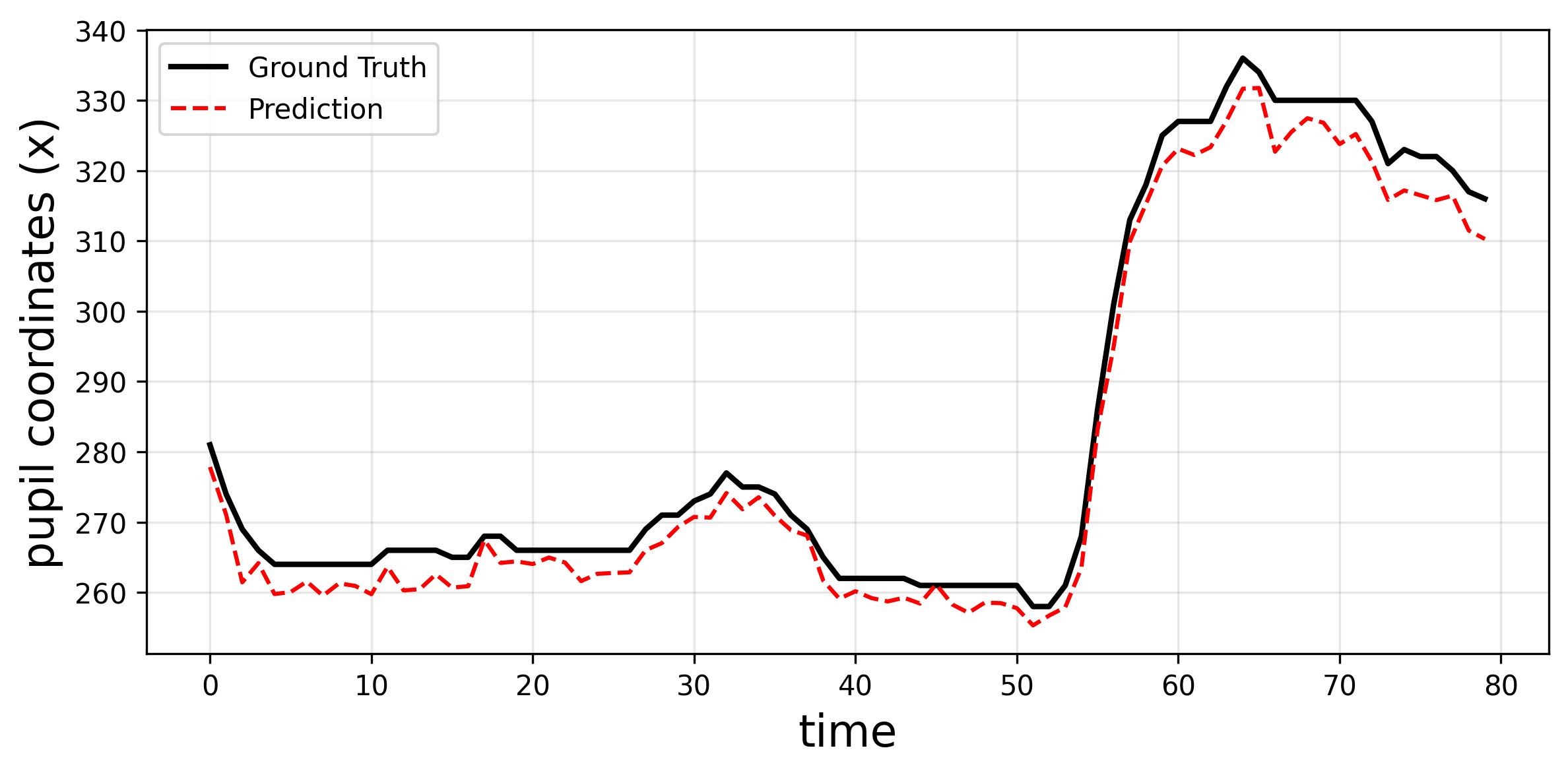}
        \caption{MSE: 14.40, JM: 0.20}
        \label{fig: (c)}
    \end{subfigure}
            \hfill
    \begin{subfigure}[t]{0.48\linewidth}
        \centering
        \includegraphics[width=\linewidth]{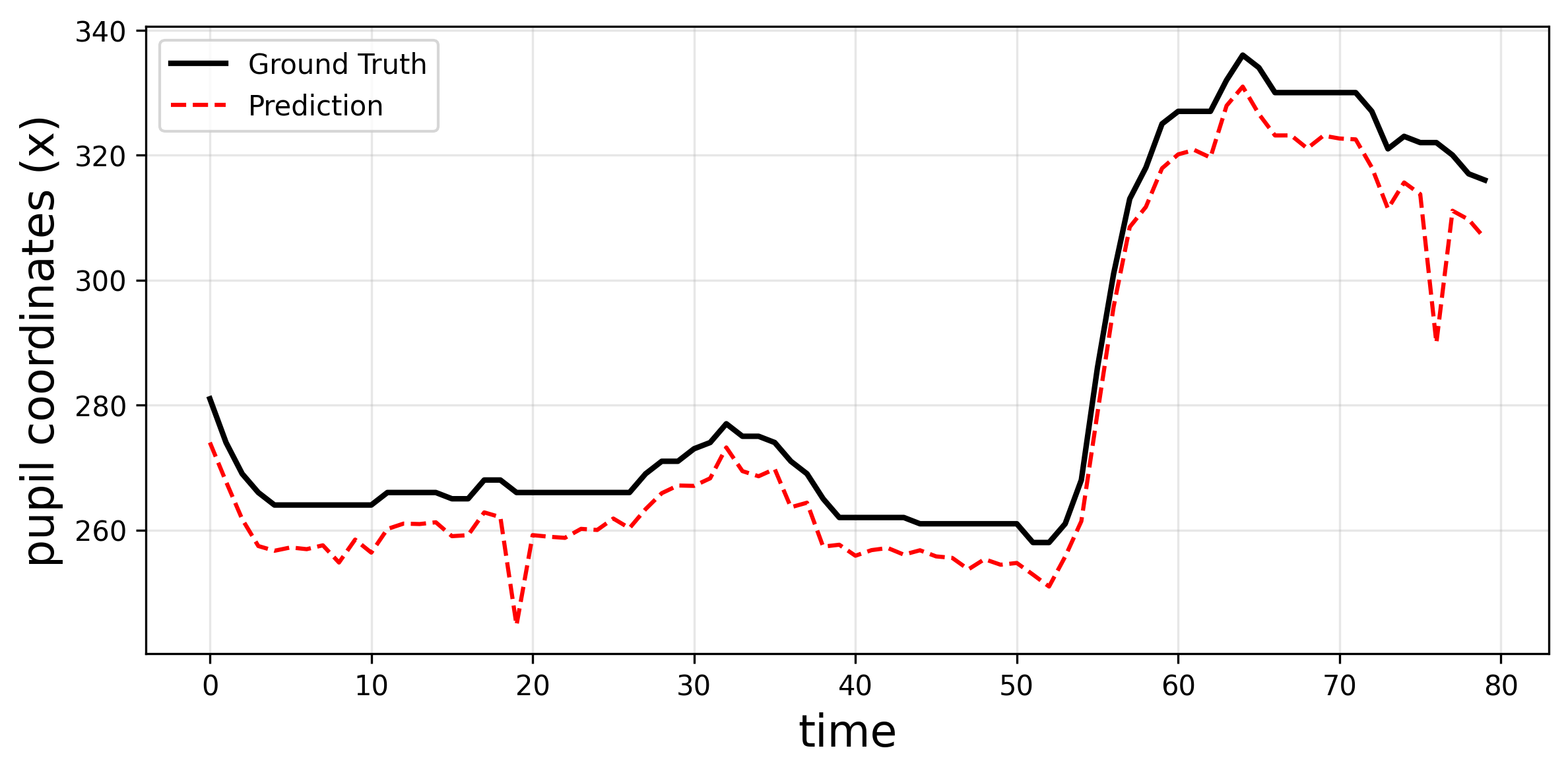}
        \caption{MSE: 59.41, JM: 0.32}
        \label{fig: (d)}
    \end{subfigure}
            \hfill
    \begin{subfigure}[t]{0.48\linewidth}
        \centering
        \includegraphics[width=\linewidth]{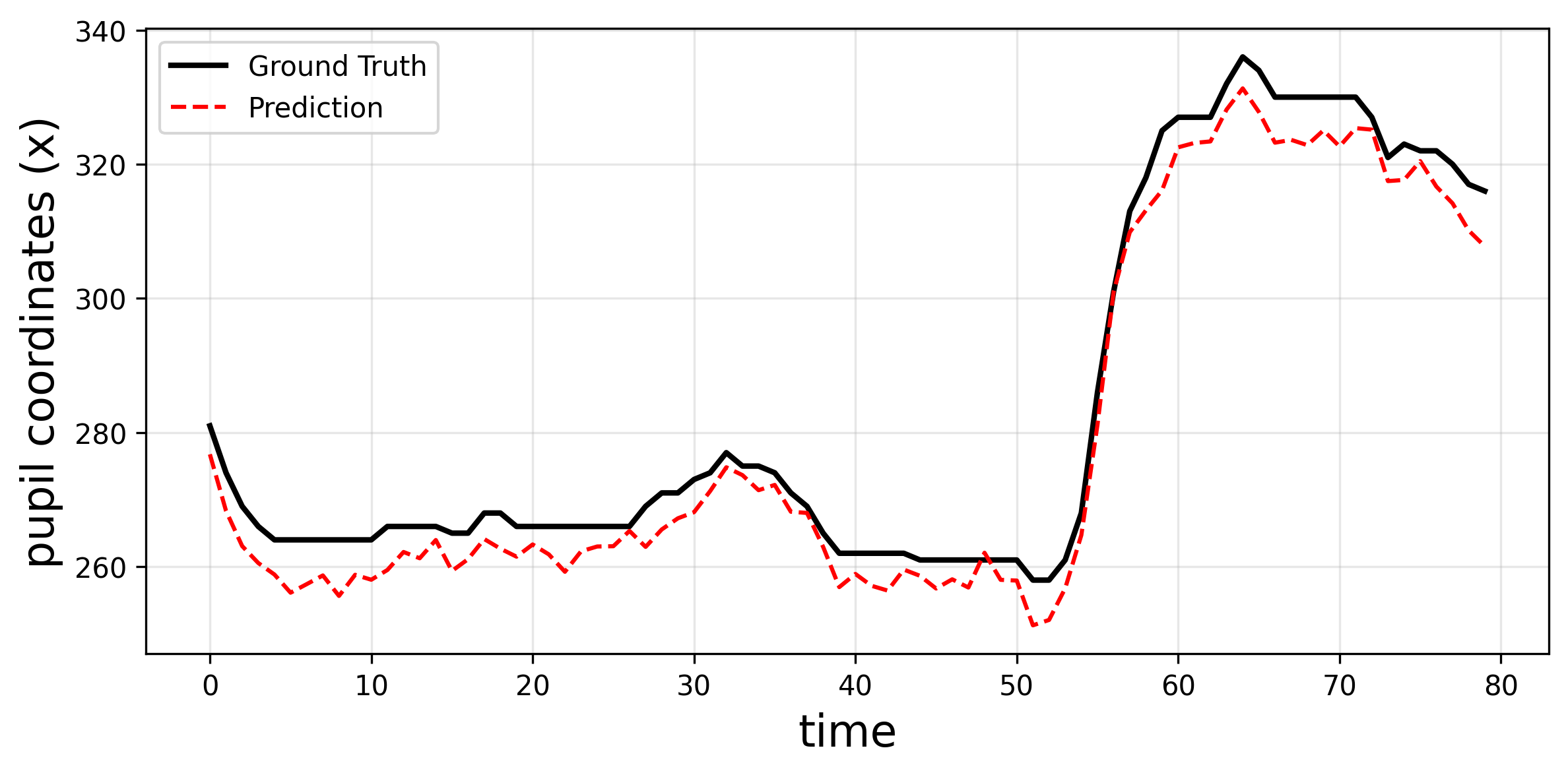}
        \caption{MSE: 23.07, JM: 0.16}
        \label{fig: (e)}
    \end{subfigure}        \hfill
    \begin{subfigure}[t]{0.48\linewidth}
        \centering
        \includegraphics[width=\linewidth]{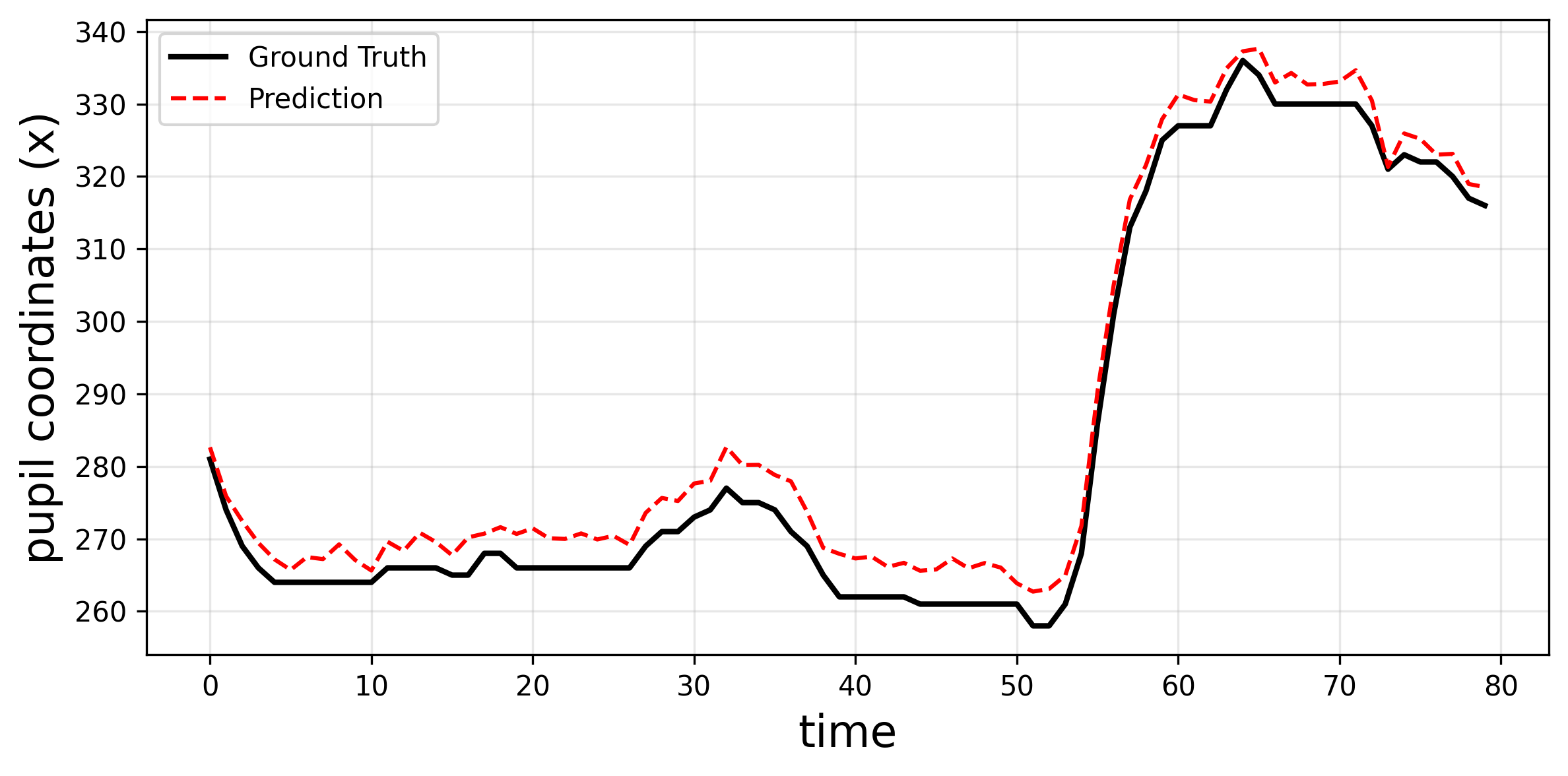}
        \caption{MSE: 16.32, JM: 0.11}
        \label{fig: (f)}
    \end{subfigure}        \hfill
    \begin{subfigure}[t]{0.48\linewidth}
        \centering
        \includegraphics[width=\linewidth]{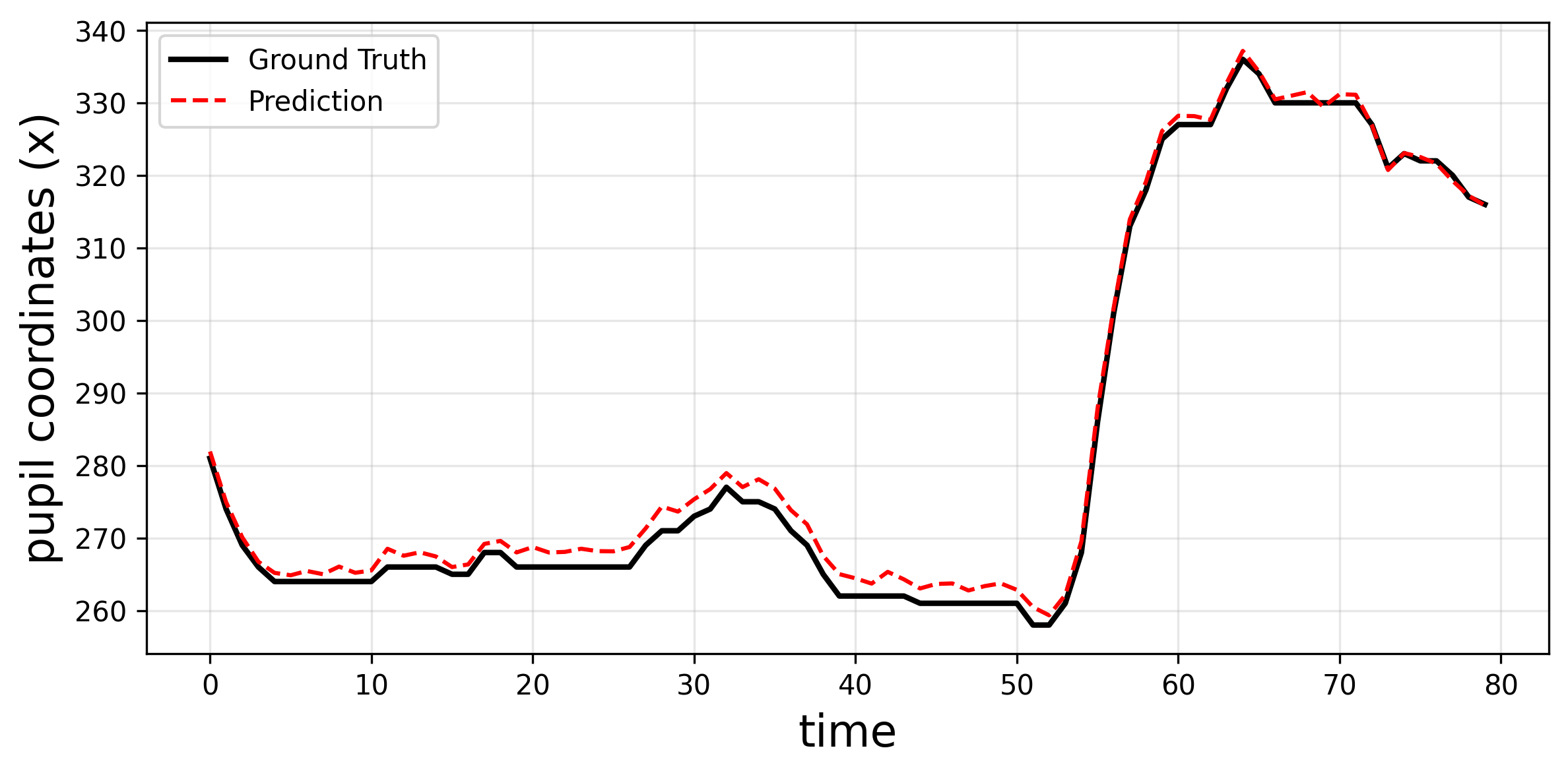}
        \caption{MSE: 3.39, JM: 0.05}
        \label{fig: (g)}
    \end{subfigure}
    \caption{Empirical (exemplary) demonstrations to show the (1) necessity of the proposed jitter metric as a complimentary metric for the task of pupil tracking and (2) efficacy of the proposed jitter metric to evaluate the temporal smooth-continuity. In all sub-figures, the x-axis represents the time domain while the y-axis represents the pupil coordinate (to be specific, x-coordinate) variation. The ground truth variation is retrieved from 3ET+ dataset and we thoughtfully perturbed the ground truth to imitate (and generate) the potential cases of predictions. The perturbation process included randomly adding (1) low-amplitude random noise, (2) blinking artifacts, (3) pixel shift artifacts, and (4) high-frequency tremor-like noise, and each generated prediction has at least two types of perturbations.}
    \label{fig: jitter_metric_empiricalDemo}
\end{figure}

To empirically demonstrate the necessity and validity of the proposed jitter metric in the context of pupil tracking, we present a set of controlled visual demonstrations in Fig.~\ref{fig: jitter_metric_empiricalDemo}. These examples are designed to contrast gaze prediction trajectories with varying degrees of temporal smoothness and positional accuracy. In doing so, we highlight the limitations of conventional metrics, such as Mean Squared Error (MSE), and illustrate how the proposed jitter metric serves as a complementary measure that captures temporal continuity, a crucial yet often overlooked dimension in eye tracking evaluation.

Each subfigure (Fig.~\ref{fig: (b)} – Fig.~\ref{fig: (g)}) represents a perturbed version of the predicted pupil trajectory (Fig.~\ref{fig: (a)}) obtained from the 3ET+ dataset. Perturbations were deliberately designed to simulate typical error modes encountered in real-world event-based eye-tracking pipelines. These include: (i) low-amplitude random noise, (ii) blink-induced discontinuities, (iii) pixel shift artifacts, and (iv) high-frequency tremor-like oscillations. Each simulated prediction includes at least two of these perturbations to reflect realistic degradation patterns observed in practice.

We use the prediction in Fig.~\ref{fig: (a)} as the reference trajectory. It achieves a moderate MSE of 15.83 and a jitter metric score of $JM_{\lambda = 0.75}$= 0.18. This example serves as a baseline for comparing other trajectories in terms of both spatial and temporal quality. In Fig.~\ref{fig: (b)}, although the MSE is lower than that of the reference, the prediction exhibits reduced temporal smoothness. This discrepancy is effectively captured by our jitter metric, which assigns it a higher score, penalizing the temporal instability overlooked by conventional positional metrics. Conversely, the prediction in Fig.\ref{fig: (e)} demonstrates superior temporal smoothness, despite a slightly higher MSE. The jitter score reflects this improvement, underscoring the metric's ability to reward smooth predictions even in the presence of minor spatial deviations.

Figures~\ref{fig: (c)} and \ref{fig: (f)} present a particularly instructive comparison: both predictions yield nearly identical MSE values but differ significantly in their temporal continuity. The proposed jitter metric distinguishes between the two, accurately assigning a lower score to the smoother prediction. These cases exemplify scenarios where traditional metrics fail to differentiate predictions with similar positional accuracy but markedly different perceptual quality.

The trajectory in Fig.~\ref{fig: (d)} is characterized by blink-related discontinuities and abrupt pixel shifts—common artifacts in event-based systems. While its MSE is comparable to the reference, its elevated jitter score reflects the increased temporal noise. This highlights the metric's sensitivity to transient disruptions that can compromise downstream tasks such as attention estimation or gaze-based interaction.

Finally, Fig.~\ref{fig: (g)} illustrates a case with relatively high MSE but excellent temporal smoothness. The jitter metric appropriately assigns it a lower score than the reference, reinforcing its utility as a decoupled measure of temporal fidelity.

These examples validate the proposed jitter metric as a critical complement to existing positional accuracy measures. By capturing trajectory-level smoothness, the metric provides a more holistic evaluation of prediction quality, particularly in applications such as micro-expression recognition, cognitive state inference, and gaze-based behavioral analytics, where temporal consistency is important.


\section{Experiments \& Results}

\subsection{Datasets \& Base Models}
\label{ssec: base_models}
By following the recent challenge on event-based eye tracking~\cite{chen2025eventvision_event}, we test our method on the 3ET+ dataset~\cite{chen2025eventvision_event, wang2024event} since 3ET+ serves as the most prominent benchmark dataset for the task at hand. In contrast, since our proposed method is presented as a post-processing step and works in a model-agnostic fashion, we select two recent models as base models: CB-ConvLSTM~\cite{chen20233et}, and bigBrains~\cite{pei2024lightweight}, to show the impact of the proposed pipeline towards improved pupil coordinate predictions in each case. More descriptively, CB-ConvLSTM is a change-based convolutional long short-term memory architecture which specifically designed for efficient spatio-temporal modelling to predict pupil coordinates from sparse event frames, whereas bigBrains attempts to preserve causality and learn spatial relationships using a lightweight model consisting of spatial and temporal convolutions.  

\subsection{Implementation Details}
We implement and run all our post-processing blocks on a single V100 GPU machine while setting the following specifics for each proposed algorithm. For the implementation of algorithm~\ref{algo:median}, we set $w_{min}$ to be $5$, $w_{max}$ to be $20$, the percentile to determine adaptive window size to be $75$ and the default mode of local motion variance estimation method to be based on covariance. For the implementation of algorithm~\ref{algo:ofe}, we set the scaling parameter $\tau$ to be $8$, the count threshold $c$ to be $5$, and the difference threshold $\gamma$ to be $2$.


\subsection{Baselines}

Trivially, we consider two base models, described in section~\ref{ssec: base_models}, as baselines to compare with the proposed post-processing techniques. In addition, to further extend our analysis, we compare our method with latest other works in the literature as well, including EyeGraph~\cite{bandara2024eyegraph}, MambaPupil~\cite{wang2024mambapupil}, FreeEVs~\cite{wang2024event}, and SEE~\cite{zhang2024co}. 

\subsection{Evaluation Metrics}


Along with the proposed jitter metric, we implement three other metrics: \textit{p-accuracy}, mean Euclidean distance ($l_2$) and mean Manhattan distance ($l_1$), which are utilized in the recent works in the literature~\cite{wang2024event}, to quantitatively evaluate the performance of the proposed post-processing methods. As defined in~\cite{chen20233et}, \textit{p-accuracy}, as presented in Eq.~\ref{eq. p-acc}, indicates the pixel-level accuracy of the predictions by checking the Euclidean distance between the predicted coordinates ($pred_i$) and true coordinates ($true_i$) is within a specified pixel threshold ($Th$). In this work, we set the pixel thresholds to be 10, 5, and 1 following~\cite{wang2024event}. Further, since the pupil coordinate prediction is a regression task, we incorporate two well-established regression metrics: $l_2$ in Eq.~\ref{eq.l_2} and $l_1$ in Eq.~\ref{eq. l_1} as well.  

\begin{equation}
\begin{split}
    p\{Th\} = \frac{1}{N}\sum_{i=1}^N f(true_i, pred_i, Th)\,\\ \text{with}\,f(true_i, pred_i, Th) = 
    \begin{cases}
    1& \text{if } \Vert true_i - pred_i \Vert \leq Th\\
    0              & \text{otherwise}
\end{cases}  
\end{split}
\label{eq. p-acc}
\end{equation}

\begin{equation}
    l_2 = \frac{1}{N}\sum_{i=1}^N \Vert true_i - pred_i \Vert_2 
    \label{eq.l_2}
\end{equation}

\begin{equation}
    l_1 = \frac{1}{N}\sum_{i=1}^N\,|\,true_i - pred_i\,| 
    \label{eq. l_1}
\end{equation}

\subsection{Results}

\begin{figure}[!h]
    \centering
    \begin{subfigure}[t]{0.24\linewidth}
        \centering
        \includegraphics[width=\linewidth]{Figures/frame_0165.png}
        \label{result_1}
    \end{subfigure}
    \hfill
    \begin{subfigure}[t]{0.24\linewidth}
        \centering
        \includegraphics[width=\linewidth]{Figures/frame_0178.png}
        \label{result_2}
    \end{subfigure}
        \hfill
    \begin{subfigure}[t]{0.24\linewidth}
        \centering
        \includegraphics[width=\linewidth]{Figures/frame_0192.png}
        \label{result_3}
    \end{subfigure}
    \hfill
    \begin{subfigure}[t]{0.24\linewidth}
        \centering
        \includegraphics[width=\linewidth]{Figures/frame_0203.png}
        \label{result_4}
    \end{subfigure}
        \begin{subfigure}[t]{0.24\linewidth}
        \centering
        \includegraphics[width=\linewidth]{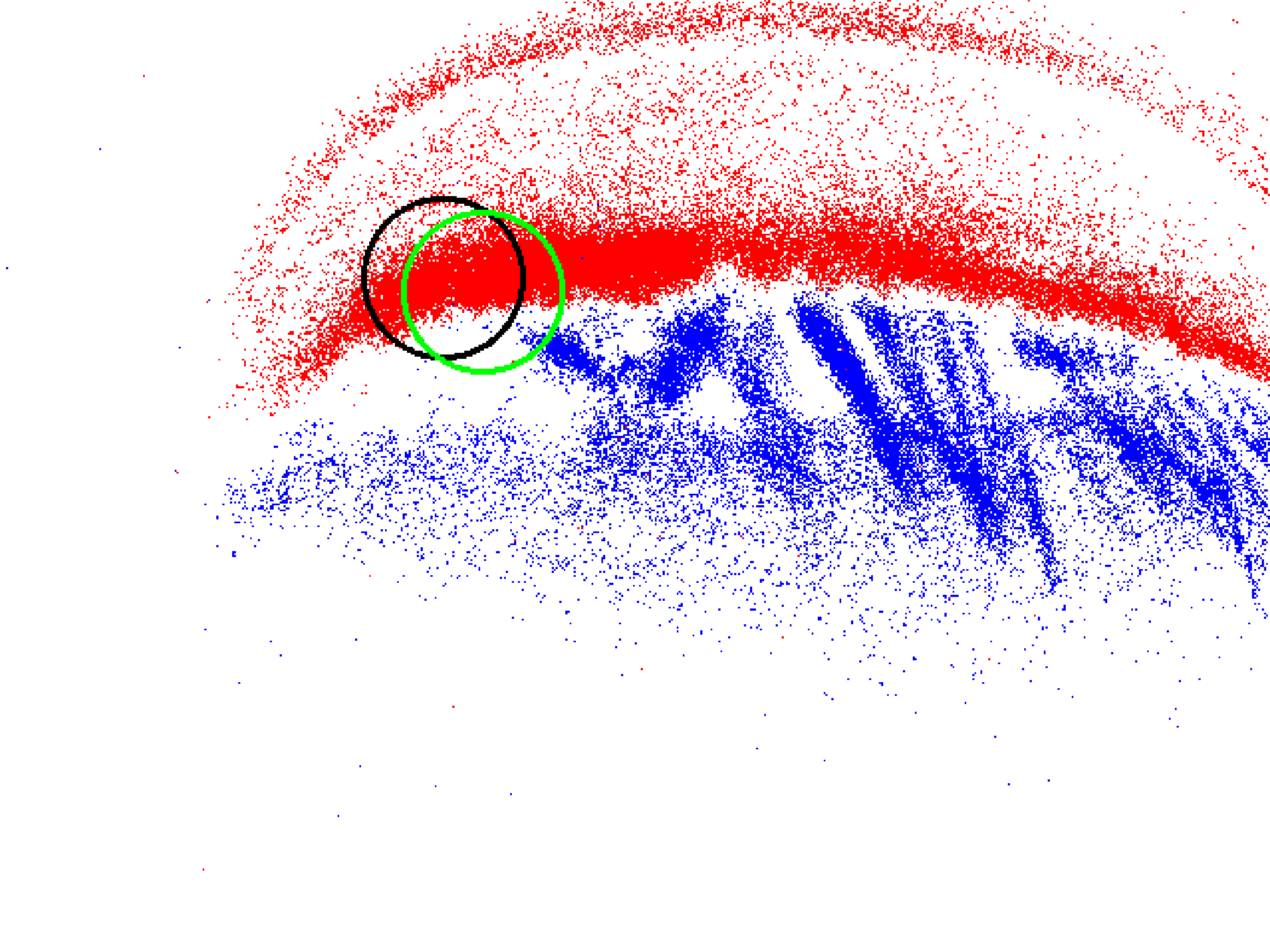}
        \label{result_5}
    \end{subfigure}
    \hfill
    \begin{subfigure}[t]{0.24\linewidth}
        \centering
        \includegraphics[width=\linewidth]{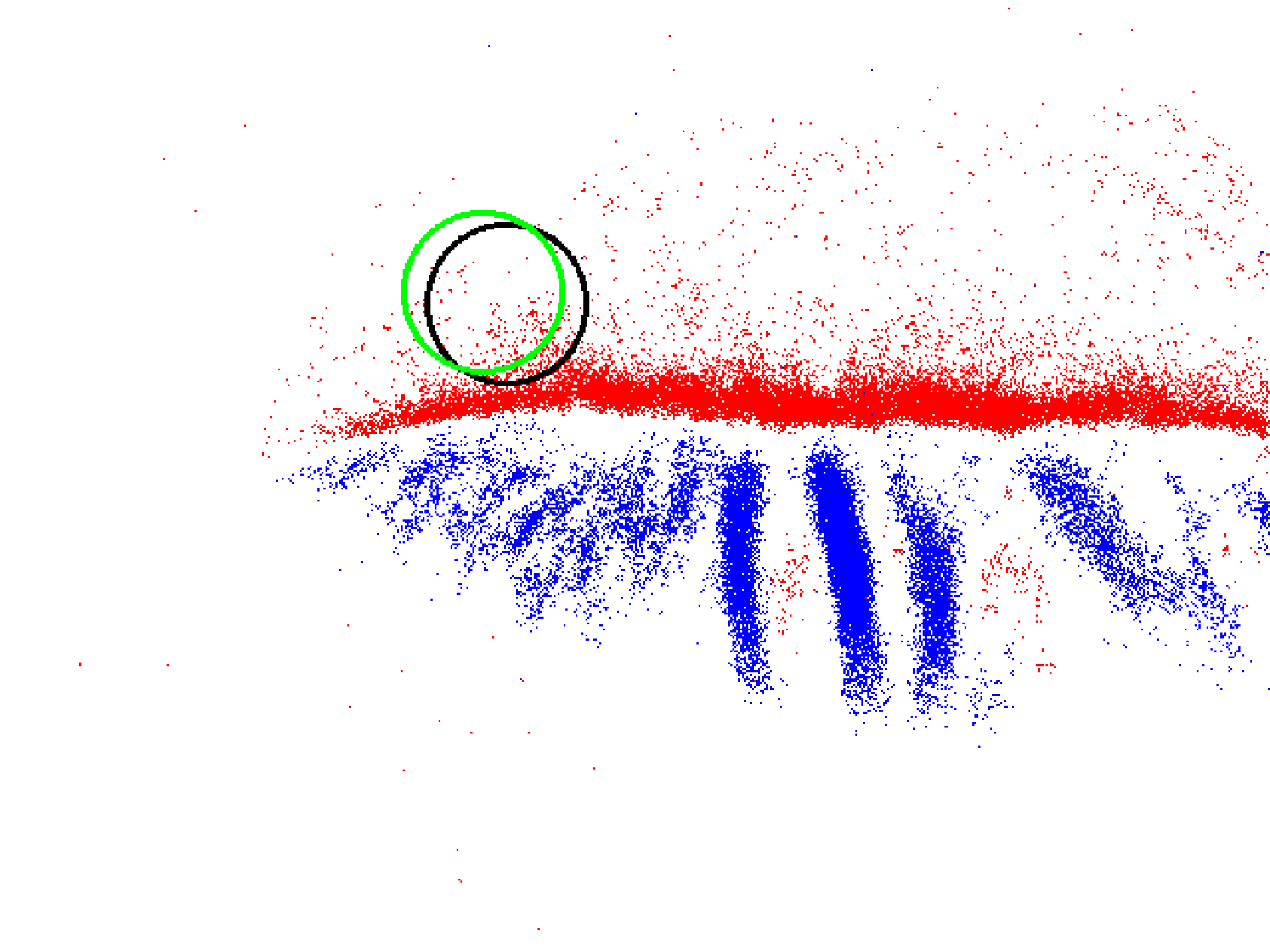}
        \label{result_6}
    \end{subfigure}
        \hfill
    \begin{subfigure}[t]{0.24\linewidth}
        \centering
        \includegraphics[width=\linewidth]{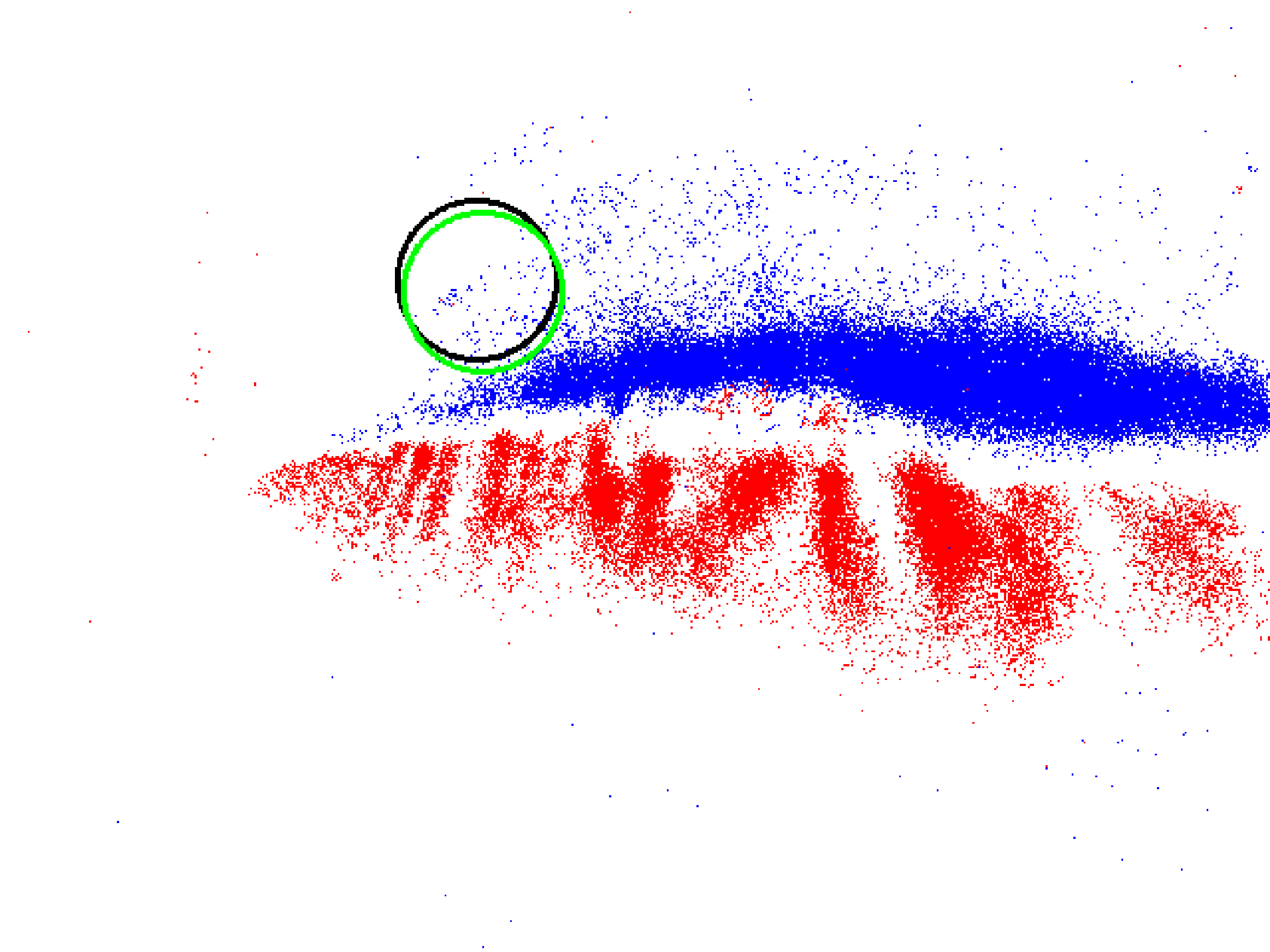}
        \label{result_7}
    \end{subfigure}
    \hfill
    \begin{subfigure}[t]{0.24\linewidth}
        \centering
        \includegraphics[width=\linewidth]{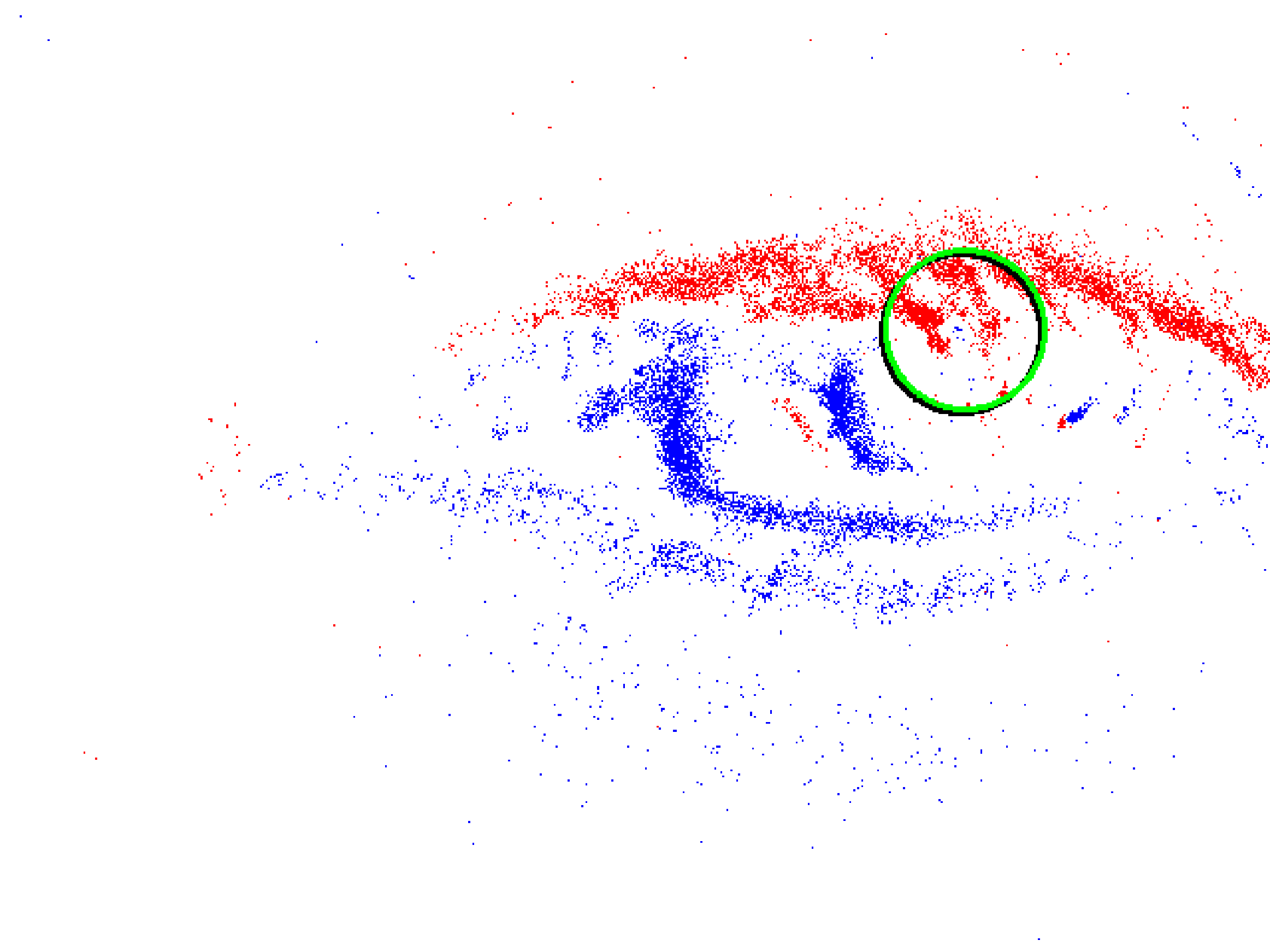}
        \label{result_8}
    \end{subfigure}
    \caption{Qualitative results. The first row present a selected set of results from~\cite{pei2024lightweight} where the second row presents the corresponding results from our methods, during a challenging pupil movement scenario. As evident, out post-processing methods improve the temporal continuity of the predictions and thereby, the overall performance of the base models.}
    \label{fig: qualitative_results}
\end{figure}

In this section, we present a comprehensive evaluation of our inference-time post-processing framework across four dimensions: standard positional accuracy metrics, our proposed jitter metric, per-component ablation analysis, and computational complexity. We also present a set of qualitative results in Fig.~\ref{fig: qualitative_results} for further demonstration. 

\textbf{Positional Accuracy:}
Table~\ref{tab:pupil_tracking} reports the performance of our inference-time refinement pipeline on standard positional accuracy metrics, evaluated on predictions from the bigBrains model~\cite{pei2024lightweight}. Our post-processing techniques consistently improve the gaze localization accuracy across all benchmarks. Notably, when applied to the base model, our method reduces the mean squared error by more than 5.1\% on average (on both validation and test datasets), outperforming all existing event-based eye tracking approaches. These results demonstrate that our method enhances baseline predictions without any model retraining or architectural modifications, validating its model-agnostic design.

\textbf{Computational Complexity:}
Table~\ref{tab:SMU_table_macs} quantifies the computational overhead introduced by our post-processing modules. As our methods operate entirely at inference time and do not include any trainable parameters, the additional computational burden is minimal. Specifically, motion-aware median filtering and optical flow refinement require approximately 172 and 340 FLOPs per event frame, respectively. Across all tested configurations, the total computational overhead remains below 0.00048\% of the base model's cost. This confirms the practicality of our approach for real-time deployment on edge devices with constrained resources.

\textbf{Temporal Smoothness via Jitter Metric:}
Table~\ref{tab:JM_table} highlights the benefits of our approach with respect to temporal smoothness, as measured by the proposed jitter metric. Unlike traditional metrics that emphasize spatial proximity to ground truth, the jitter metric captures the fine-grained continuity of gaze predictions over time, an essential attribute for downstream applications such as mind-state decoding or attention estimation. Our refinement modules yield significantly lower jitter scores than both the base model and other state-of-the-art systems, indicating smoother and more stable tracking output. This demonstrates that our method not only improves accuracy but also mitigates high-frequency noise often induced by sensor sparsity or blinking.

\textbf{Ablation Analysis:}
To isolate the contribution of each refinement component, we conduct an ablation study summarized in Table~\ref{tab:SMU_table}. We evaluate the individual effects of the motion-aware median filtering and optical flow-based local refinement modules. The former proves effective in suppressing transient spikes caused by blinks and sensor noise, while the latter aligns predictions more closely with the underlying motion cues embedded in the event stream. When combined, these components exhibit complementary benefits, leading to the highest overall improvements in both accuracy and smoothness. These results validate the composability and robustness of our design.

\begin{table}[!t]
    \centering
    \begin{tabular}{llllllll}
        \hline
        Method  & Supervised? & Post-processing?$^\dagger$ & p10$\uparrow$  & p5$\uparrow$ & p1$\uparrow$  & $l_2$$\downarrow$  & $l_1$$\downarrow$ \\
        \hline
        EyeGraph~\cite{bandara2024eyegraph}   & \xmark & \xmark & 91.45 & 89.22 & 28.34 & 3.88 & 4.24\\
                \hline
        MambaPupil~\cite{wang2024mambapupil}    & \cmark & \xmark &\underline{99.42} & 97.05 & 33.75 &1.67 &2.11\\
        \hline
        FreeEVs~\cite{wang2024event}    &  \cmark & \xmark& 99.26 & 96.31 & 23.91 & 2.03 & 2.56\\
        \hline
        bigBrains~\cite{pei2024lightweight}    & \cmark & \xmark& 99.00 & \underline{97.79} & \underline{45.50} & \underline{1.44} & \underline{1.82}\\
        \hline
        SEE~\cite{zhang2024co}    & \cmark & \xmark& 99.00 & 77.20 & 7.32 & 3.51 & 4.63\\
                \hline
        Ours (based on~\cite{pei2024lightweight}) & \cmark & \cmark& \textbf{99.99}& \textbf{99.84}& \textbf{59.87}& \textbf{0.80} & \textbf{0.65}\\
        \hline
    \end{tabular}
    \caption{Pupil tracking results on 3ET+ validation dataset. $^\dagger$Here, we define whether post-refinement is applied to the original predictions to get the final predictions. The results of the prior works are extracted from the respective papers~\cite{bandara2024eyegraph, wang2024event}.}
    \label{tab:pupil_tracking}
\end{table}

\begin{table}[!t]
    \centering
    \begin{tabular}{c|c|c|c|c}
        \hline      
        Method &  M2F$^\ddagger$ & OFE$^\ddagger$ & Model size& $\#$ MACs\\
        \hline
        ConvGRU~\cite{chen20233et} & \xmark & \xmark &  417K & 2716.419840M \\
        ConvGRU~\cite{chen20233et} & \cmark & \cmark &  417K & 2716.420096M \\
        bigBrains~\cite{pei2024lightweight} & \xmark & \xmark & 809K &  59.537664M \\
        bigBrains~\cite{pei2024lightweight} & \cmark & \cmark & 809K & 59.537920M \\
        \hline
    \end{tabular}
    \caption{Computational complexity details of the proposed modules. $^\ddagger$Here we refer to our motion-aware median filtering and optical flow-based local refinement as M2F and OFE respectively.}
    \label{tab:SMU_table_macs}
\end{table}

\begin{table}[!t]
    \centering
    \begin{tabular}{c|c|c|c}
        \hline      
        Method &  M2F$^\ddagger$ & OFE$^\ddagger$ & JM$\downarrow$\\
        \hline
        bigBrains~\cite{pei2024lightweight} & \xmark & \xmark & 0.4936 \\
        bigBrains~\cite{pei2024lightweight} & \cmark & \cmark &  \textbf{0.4372} \\
        \hline
    \end{tabular}
    \caption{Results comparison with respect to the proposed jitter metric on 3ET+ validation dataset. $^\ddagger$Here we refer to our motion-aware median filtering and optical flow-based local refinement as M2F and OFE respectively.}
    \label{tab:JM_table}
\end{table}

As shown in Tab.~\ref{tab:SMU_table}, both of our post processing techniques consistently improved the results of vanilla predictions of each method. To this end, only applying the motion-aware median filtering improves the vanilla prediction performance of~\cite{pei2024lightweight} by reducing the $l_2$ error from $1.500$ to $1.466$ whereas applying both motion-aware median filtering and optical flow-based local refinement leads to an $l_2$ of $1.423$, thereby marking an overall improvement of $5.13\%$. Similarly, when we apply our model-agnostic methods on~\cite{chen20233et}'s vanilla predictions, the performance improved from $7.922$ to $7.504$. These observations confirm the validity of the proposed model-agnostic post-processing methods as a collective way of improving the existing models while also ensuring the efficacy of the individual blocks. 

\begin{table}[!t]
    \centering
    \begin{tabular}{c|c|c|c|c}
            \hline
        Method &  M2F$^\ddagger$ & OFE$^\ddagger$ & $l_2$$\downarrow$ & $l_2$$\downarrow$ \\
         &   &  & (Public) & (Private)\\
        \hline
        ConvGRU~\cite{chen20233et} & \xmark & \xmark & 7.914 & 7.922\\
        ConvGRU~\cite{chen20233et} & \cmark & \cmark & 7.494 & 7.504\\
        bigBrains~\cite{pei2024lightweight} & \xmark & \xmark & 1.431 & 1.500 \\
        bigBrains~\cite{pei2024lightweight} & \cmark & \xmark & 1.408 & 1.466 \\
        bigBrains~\cite{pei2024lightweight} & \cmark & \cmark & \textbf{1.382} & \textbf{1.423}\\
        \hline
    \end{tabular}
    \caption{Ablations to show the performance improvement by the proposed modules. The results are from the test set at the eye tracking challenge~\cite{chen2025eventvision_event}. $^\ddagger$Here we refer to our motion-aware median filtering and optical flow-based local refinement as M2F and OFE respectively.}
    \label{tab:SMU_table}
\end{table}

\section{Discussion \& Conclusion}
\label{sec: discussion}

This work presents a practical, model-agnostic framework for enhancing event-based eye-tracking pipelines at inference time. A key strength of our approach is its applicability: the proposed post-processing techniques, i.e., Motion-Aware Median Filtering and Optical Flow Estimation for smooth shifts, can be seamlessly applied to the output of any existing event-based pupil estimation model, improving temporal stability and spatial coherence without retraining or modifying the model architecture. This makes our method especially valuable in resource-constrained settings or when dealing with black-box models, offering a lightweight way to boost performance across a wide range of real-world applications. Additionally, the introduction of a dedicated Jitter Metric provides a complementary measure on model quality, addressing a critical gap in evaluation criteria for time-sensitive behavioral tracking.

Despite these strengths, there are important limitations to acknowledge.
\begin{itemize}
    \item Our work currently focuses exclusively on the ocular modality. While eye movements, particularly micro-saccades and pupil dynamics, are powerful indicators of cognitive states such as attention, fatigue, or confusion, real-world mind-state inference typically benefits from multi-modal integration, combining gaze with facial micro-expressions, head dynamics, or physiological signals. Extending our refinement pipeline and temporal metrics to accommodate or complement such modalities remains an exciting direction for future work.

    \item Our evaluations are conducted on datasets collected in controlled laboratory settings, where participants are relatively still, lighting is consistent, and noise in the event stream is minimal. In contrast, real-world deployments, for example, in wearable settings or during naturalistic interactions, introduce challenges such as head motion, background clutter, dynamic lighting, and partial occlusions. These conditions may degrade the assumptions behind our refinement techniques (e.g., motion coherence in flow estimation), and thus real-world validation is a critical next step.

    \item While our approach improves temporal smoothness and reduces spatial jitter, it does not correct fundamental prediction errors arising from poor base model performance. If a baseline model consistently mispredicts pupil location due to sensor misalignment, incorrect calibration, or biased training data, our method may smooth those errors rather than eliminate them. Therefore, the method is best viewed as an enhancement layer for models that already offer reasonable accuracy, rather than as a full corrective mechanism.
\end{itemize}

\begin{acknowledgments}
  This work was supported by both the Ministry of Education (MOE) Academic Research Fund (AcRF) Tier 1 grant (Grant ID: 22-SIS-SMU-044), and by Singapore Management University’s Lee Kong Chian Professorship Award.  Any opinions, findings and conclusions or recommendations expressed in this material are those of the author(s).
\end{acknowledgments}

\section*{Declaration on Generative AI}
  
 During the preparation of this work, the author(s) used ChatGPT in order to: Grammar and spelling check. After using this tool, the author(s) reviewed and edited the content as needed and take(s) full responsibility for the publication’s content. 

\bibliography{main}





\end{document}